\documentclass[11pt]{article}

\usepackage[final]{acl}

\usepackage{times}
\usepackage{latexsym}
\usepackage{bbding}
\usepackage[T1]{fontenc}
\usepackage[utf8]{inputenc}
\usepackage{etoc}
\usepackage{microtype}

\usepackage{inconsolata}

\usepackage{graphicx}

\usepackage{hyperref}
\usepackage{url}
\usepackage{booktabs}
\usepackage{makecell}
\usepackage{graphicx} 
\usepackage{multirow}
\usepackage{graphicx}
\usepackage{subcaption}
\usepackage{wrapfig} % 需要添加宏包
\usepackage{booktabs} % For better-looking tables
\usepackage[table]{xcolor} % For cell coloring
\usepackage{wrapfig}
\usepackage{xurl}
\usepackage{amsmath}
\usepackage{ragged2e}
\usepackage{subcaption}
\usepackage[export]{adjustbox} % for valign=t
 \usepackage{algorithm,algpseudocode}
\usepackage{amssymb} 
\definecolor{textblue}{RGB}{70,130,180}
\definecolor{textred}{RGB}{205,92,92}
\definecolor{lightblue}{rgb}{0.8, 0.9, 1}
\definecolor{lightpink}{RGB}{255,192,203}
\definecolor{lightblue}{rgb}{0.8, 0.9, 1}
\definecolor{lightyellow}{RGB}{255,255,224}
\definecolor{lightgreen}{RGB}{144,238,144}
\definecolor{gray}{RGB}{220 220 220}
\usepackage{enumitem}
\usepackage[nottoc]{tocbibind}

\usepackage{array}
\usepackage{arydshln}
\newcommand{\capcolorbox}[2]{%
  {\setlength{\fboxsep}{0.6pt}\colorbox{#1}{#2}}%
}
\newcommand{\lightsep}{%
  \arrayrulecolor{gray}
  \specialrule{0.3pt}{1pt}{1pt}
  \arrayrulecolor{black}
}
\title{To Persist or To Rectify? Unveiling LLMs' Fact-Checking Behavior Under Discrepancy between Internal Prior and Retrieved Context}
\title{PAVE: A Diagnostic Testbed for Model-Conditioned Veracity Arbitration in LLM Fact-Checking}
\title{Diagnosing LLM Arbitration Behavior over Pre-evidence Epistemic States in RAG-based Fact-Checking}

\author{
  Yuxi Sun\textsuperscript{1} \quad 
  Wenbo Shang\textsuperscript{1} \quad  
  Wei Gao\textsuperscript{2} \quad 
  Xin Huang\textsuperscript{1} \quad 
  Jing Ma\textsuperscript{1}\thanks{Corresponding author.} \\
  \textsuperscript{1}Hong Kong Baptist University 
  \textsuperscript{2}Singapore Management University \\
  {\texttt{\{csyxsun, cswbshang, jingma,  xinhuang\}@comp.hkbu.edu.hk}}\ 
   { \texttt{weigao@smu.edu.sg}} 
}

\begin{document}
\maketitle
\begin{abstract}
In RAG-based fact-checking, LLMs are increasingly used as verifiers to check given claims against retrieved evidence. Their parametric knowledge can induce pre-evidence tendencies that may conflict with the retrieved context, yet existing evaluation frameworks do not characterize such prior-context discrepancy or measure how verifiers arbitrate between parametric and contextual signals. We introduce \textsc{PAVE} (\emph{Prior-Aware Verifier Evaluation}), a diagnostic testbed that stratifies an LLM verifier into four epistemic states based on the correctness and confidence of its pre-evidence prior and evaluates its arbitration behavior on this new benchmark, i.e., whether it persists in correct prior under misleading evidence, and whether it corrects wrong prior when accurate evidence is provided. Experiments across seven LLMs reveal unreliable and highly model-dependent prior-context arbitration, highlighting the importance of verifier selection for real-world RAG-based fact-checking applications. 
Based on these findings, we propose a lightweight JSD-based test-time arbitration method that improves factual reliability without modifying the underlying model, achieving competitive performance across diverse LLM families.
\end{abstract}

\section{Introduction}
 With the rapid development of large language models (LLMs), which encode broad world knowledge and support reasoning over complex claims and evidence~\citep{brown2020language,ouyang2022training}, LLM-based fact-checking has become increasingly prominent. Existing systems broadly follow two routes: judging claims using the model's parametric knowledge without explicit retrieval~\citep{jiang-etal-2020-know,roberts-etal-2020-much,zhao2023retrieving}, or augmenting the model with retrieved evidence in a RAG framework to compensate for static, incomplete, or outdated parametric memory~\citep{qin2024tool,shi2024replug,pan2023fact}.
\begin{figure}
    \centering
    \includegraphics[width=1\linewidth]{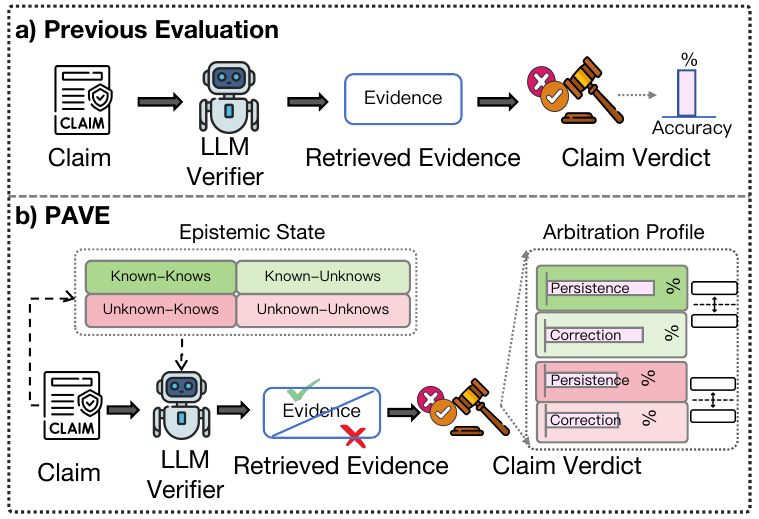}
    \caption{\textbf{Overview of \textsc{PAVE}.} Conventional evaluation judges verifiers by final-verdict accuracy on retrieved evidence (a). \textsc{PAVE} (b) instead characterizes verifiers by both their pre-retrieval \emph{epistemic state} (four Knowledge-Boundary categories) and their \emph{arbitration profile} under prior-context discrepancy (persistence \& correction).}
    \label{fig:placeholder}
    \vspace{-0.5em}
\end{figure}

Yet RAG-based verification remains shaped by the interaction between an LLM's parametric knowledge and retrieved external context. This raises two evaluation challenges. \textbf{First}, when prior knowledge disagrees with retrieved context, as can occur with noisy retrieval or temporally outdated knowledge~\citep{shi2023large,luu2022time,xie2023adaptive}, the verifier must arbitrate between parametric and contextual signals. We term this condition \emph{prior-context discrepancy} (PCD), an important but underexplored setting in real-world fact-checking. Existing evaluation pipelines are not designed to handle PCD: standard fact-checking \textit{benchmarks}~\citep{thorne2018fever, aly2021feverous,anand2024quantemp,ma2024ex} pair claims with clean gold-standard evidence, structurally precluding any discrepancy; and common evaluation protocols focus on final verdict correctness, leaving how the verifier arbitrates between prior knowledge and retrieved evidence unexamined. \textbf{Second}, such arbitration is conditioned by the verifier's pre-retrieval epistemic state with respect to the claim. A model that is confidently wrong, uncertain, or already correct may respond differently to the same retrieved evidence, yet these distinctions are collapsed under standard outcome-level evaluation.

We address these challenges with \textsc{PAVE} (\textbf{\underline{P}}rior-\textbf{\underline{A}}ware \textbf{\underline{V}}erifier \textbf{\underline{E}}valuation), a diagnostic testbed that evaluates how LLM verifiers arbitrate between parametric knowledge and retrieved evidence under PCD in fact-checking. \textbf{First}, we construct a PCD benchmark of over 7,000 claims under controlled counterfactual and temporal conflict scenarios, enabling systematic evaluation of verifier behavior when parametric and contextual signals disagree.
\textbf{Second}, to support state-conditioned evaluation, we characterize each verifier's pre-evidence epistemic state via a 
Knowledge Boundary (KB) framework, building on prior work on 
LLM knowledge awareness~\citep{kadavath2022language, yin-etal-2024-benchmarking}.
We operationalize KB along two observable axes: \textit{factual correctness},
indicating whether the model's prior-only prediction aligns with ground truth,
and \textit{parametric confidence}, estimated by prediction consistency 
across repeated stochastic generations as a proxy for epistemic 
stability~\citep{chen-mueller-2024-quantifying, farquhar2024a, chen2025teaching}.
% inspired by Knowledge Boundary (KB)~\citep{kadavath2022language, yin-etal-2024-benchmarking,amayuelas2024knowledge}, an operational characterization of the  verifier's
% pre-evidence epistemic state. We adopt and operationalize KB along two observable axes: \textit{factual correctness}, indicating whether the model's prior-only prediction aligns
% with ground truth, and \textit{parametric confidence}, estimated by prediction
% consistency across repeated stochastic generations as a proxy for
% epistemic stability~\citep{chen-mueller-2024-quantifying, farquhar2024a, chen2025teaching}.
Their combination yields four KB conditions:
\textit{Known-Knows} (confident and correct),
\textit{Known-Unknows} (confident but incorrect),
\textit{Unknown-Knows} (uncertain but correct), and
\textit{Unknown-Unknows} (uncertain and incorrect).
This distinction is crucial because the same verdict-level failure
% \textcolor{purple}{how about change " verdict-level failure" to "final verdict outcome"?}\wg{no need as you're talking about incorrect predictions} 
can arise from different
underlying conditions, e.g., a confidently wrong verifier that resists sound evidence differs fundamentally from an uncertain verifier that lacks a stable basis
for judgment. Together, the PCD benchmark and KB stratification produce arbitration profiles
for individual LLM verifiers, revealing dimensions of factual reliability that
remain hidden under final verdict evaluation, as illustrated in Figure~\ref{fig:placeholder}.

Evaluating seven LLMs, we find that the behavioral gap across models is substantial: Deepseek-v3 achieves a favorable balance between correction and persistence, while Llama3-8B is far more likely to be misled by counterfactual context than corrected by accurate evidence. Within the same model family, larger models demonstrate improved arbitration behavior, suggesting that scaling contributes to more reliable prior-context arbitration. We further find that stable errors can persist even when accurate evidence is available, and that models are generally more receptive to acquiring missing or outdated knowledge than to revising confident yet incorrect priors. These findings demonstrate that prior-context arbitration behavior is a systematic and measurable property of LLM verifiers that only final accuracy cannot capture.
% \wg{unclear what veracity-level means. $\rightarrow$verdict-level evaluation? It seems your measure are all based on verdict prediction} cannot capture.} 
Motivated by these findings, we propose a lightweight JSD-based test-time arbitration method that improves factual reliability without retraining or modifying the underlying model, achieving best or competitive performance in five out of six evaluated LLM families\footnote{{Our datasets are released at \url{https://doi.org/10.5281/zenodo.18151788}}}. 

Our main contributions are fourfold:
% \vspace{-0.5em}
\begin{itemize}[leftmargin=*]
% \vspace{-0.3em}
    \item We formulate LLM-based fact-checking under PCD as an epistemic conditioned arbitration problem, shifting evaluation from final verdict accuracy to how verifiers balance internal priors and external evidence under source discrepancy.
    % \vspace{-0.8em}
    \item We introduce a diagnostic testbed \textsc{PAVE}, including a controlled benchmark
    that induces PCD through counterfactual and temporally novel scenarios,
    KB stratification for characterizing pre-evidence
    epistemic states, and state-conditioned metrics that measure how verifiers resist, revise, or over-follow contextual evidence.
    % \vspace{-0.8em}
    \item Based on PAVE, we systematically evaluate seven
    LLMs under counterfactual and temporally novel scenarios,
    revealing distinct arbitration profiles and offering practical
    guidance for model selection in RAG-based fact-checking.
    % \vspace{-0.8em}
    \item We propose a lightweight JSD-based
    test-time arbitration method that improves factual reliability
    without changing the underlying model.
\end{itemize}

\section{Related Work}

\paragraph{LLM-based fact-checking.}
LLMs have demonstrated strong capabilities in knowledge-intensive
reasoning~\citep{sun2025causalabstain, yu2023kola, kong2025sharp}, and have become widely adopted as
automated fact-checking
verifiers~\citep{zhang-gao-2023-towards, tang-etal-2024-minicheck,
wang2024explainable, ortu-etal-2024-competition, mahaut2024factual,
kong2026reflex}. Such systems either rely directly on parametric
knowledge~\citep{jiang-etal-2020-know, roberts-etal-2020-much,
zhao2023retrieving} or augment the model with retrieved evidence
under a RAG framework to compensate for outdated or incomplete
parametric memory~\citep{zhang-etal-2024-retrievalqa, qin2024tool,shi2024replug, borgeaud2022improving, ram2023context,
chang-etal-2025-main}. In both cases, evaluation has largely centered on
final verdict accuracy, leaving how parametric beliefs interact with retrieved evidence under source conflict underexamined.

\paragraph{Prior-context discrepancy in RAG.}
Retrieved evidence may be noisy, outdated, or inconsistent with an LLM's parametric knowledge~\citep{xie2023adaptive, wu2024clasheval, huang2024trust, zhang-gao-2024-reinforcement, wang-etal-2025-astute}.
Prior work has studied such discrepancies through controlled perturbations, including entity substitution~\citep{longpre-etal-2021-entity, wu2024clasheval}, negation injection~\citep{zhou2023context}, and LLM-generated counterfactuals~\citep{xie2023adaptive}, showing that models vary in their susceptibility to conflicting evidence and that confidence can affect whether retrieved content overrides parametric knowledge~\citep{wu2024clasheval}. However, these studies largely measure susceptibility in aggregate, without conditioning failures on the verifier's epistemic state. Consequently, they cannot distinguish when a model appropriately persists in a reliable prior, corrects an unreliable one, or fails by over-trusting either source. We address this gap with an epistemic-state-conditioned evaluation protocol and arbitration profiles that diagnose model-specific behavior under PCD.

\paragraph{Knowledge boundary and uncertainty estimation.}
Prior work has studied LLM knowledge boundaries~\citep{garner1989metacognition,yin-etal-2024-benchmarking} by benchmarking how well models distinguish facts they are confident about from those they are uncertain or wrong~\citep{kadavath2022language,yin-etal-2024-benchmarking,sun2026fact}, and by examining models' ability to recognize uncertainty over questions with no definitive answers~\citep{amayuelas2024knowledge, li2024knowledge}.
% While these frameworks have informed studies of self-knowledge in question answering~\citep{chen2025teaching}, they have not been operationalized as evaluation protocols for fact-checking underPCD. 
For estimating parametric confidence without model internals, output-side consistency methods use prediction variability across repeated generations as a proxy for epistemic uncertainty~\citep{taubenfeld2025confidence,chen-mueller-2024-quantifying}.
Building on these ideas, we formulate KB as an observable characterization of a verifier's pre-evidence epistemic state, jointly capturing prior-only factual correctness and parametric confidence. This state characterization supports both KB-conditioned arbitration evaluation and the JSD-based test-time arbitration signal used in our method.

\begin{figure*}
    \centering
    \includegraphics[width=0.98\linewidth]{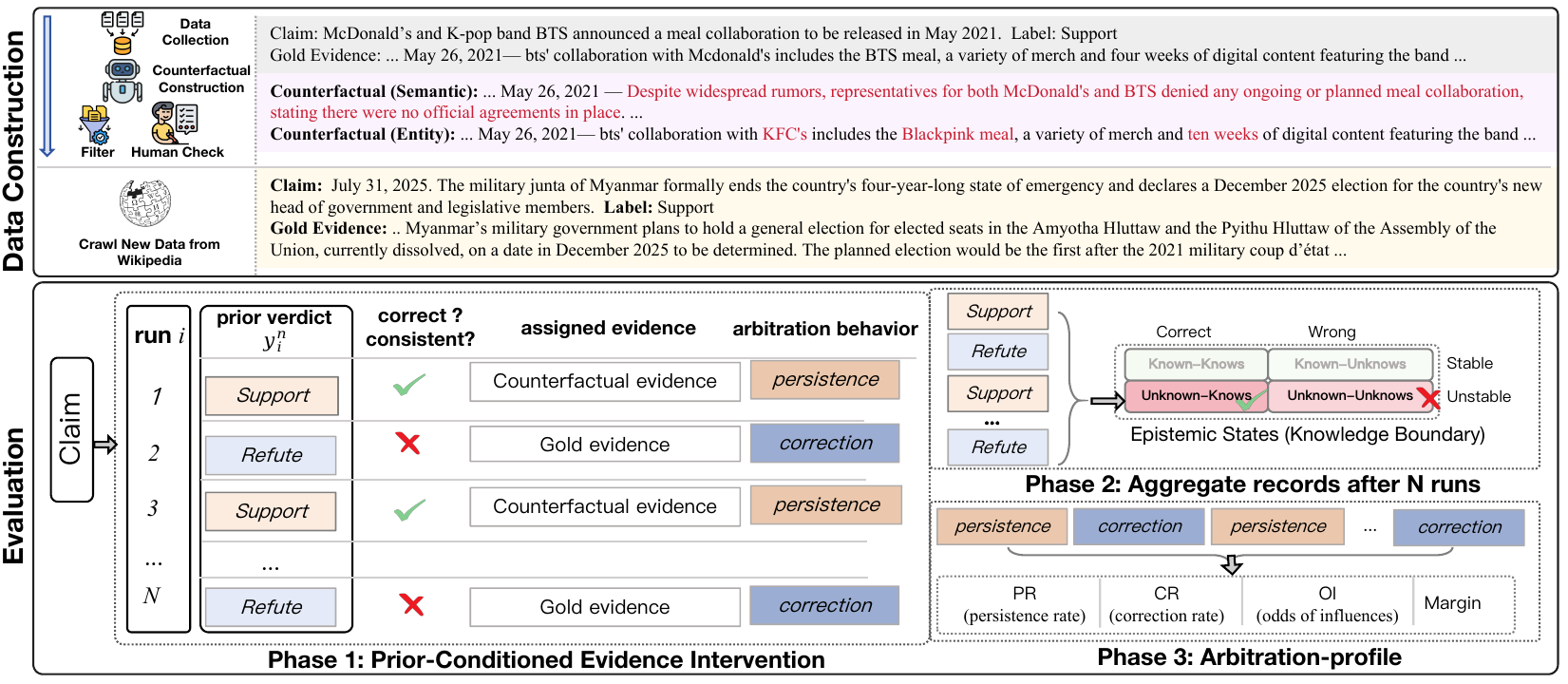}
    % \vspace{-2.2em}
    \caption{Dataset Construction and evaluation pipeline for model behavior analysis under epistemic states.}
    % \wg{Are the epistemic states marked correctly? Stable corresponds to Known and Correct corresponds to Knows, but why Known-Unknows is Unstable intersecting Correct?}}
    % \textcolor{red}{the left part of the figure is not referred in the text?}}
    % \textcolor{red}{Dataset name TWINE is not defined and used in the text. If you're keen to the name, make sure it's meaningfully defined.}}
    \label{fig:figure2}
    \vspace{-1em}
\end{figure*}

\section{Evaluation Method}

We adopt a consistency-based probing setup to characterize a verifier's pre-evidence epistemic state. For each claim $\mathbf{c}_i$ with ground-truth label $y_i$, we query the model \emph{without} retrieved evidence for $N$ independent runs, obtaining $N$ prior-only verdicts. 
%$\mathcal{V}_i=\{v_i^{(1)},\ldots,v_i^{(N)}\}$. 
\emph{Confidence} is defined by strict consistency: a claim is \textsc{Known} if all prior-only verdicts are identical; otherwise it is \textsc{Unknown}. \emph{Correctness} defines \textsc{Knows}/\textsc{Unknows}: a verdict is \textsc{Knows} if it matches $y_i$; otherwise it is \textsc{Unknows}. Combining the two axes yields four observable states for each prediction: \textsc{Known-Knows} (confident and correct), \textsc{Known-Unknows} (confident but incorrect), \textsc{Unknown-Knows} (uncertain but correct), and \textsc{Unknown-Unknows} (uncertain and incorrect), as shown in Figure~\ref{fig:figure2} (bottom).

\subsection{Evaluation Metrics}
\label{sec:metrics}
% \wbshang{the connection of metric and KB}
To comprehensively quantify the dynamics between internal prior and retrieved context, PAVE adopts a hierarchical evaluation framework. Building on the defined \textit{correction} and \textit{persistence} scenarios, we report two primary performance rates, \textbf{Correction Rate} (CR) and \textbf{Persistence Rate} (PR), which capture a model’s adaptability and robustness, respectively. We further introduce derived metrics, \textbf{Margin} and \textbf{Odds of Influence}, to analyze the trade-off between these competing behaviors. 
% For the acquisition scenario, we use \textbf{AR} to measure, which is the same calculation as CR.
% \wbshang{[Acquisition metric]}

\paragraph{Primary Metrics (CR \& PR).}
We first stratify evaluation based on the model's prior-only prediction correctness (i.e., \textsc{Knows vs. Unknows}). Consider a fact-checking test set $\{(\mathbf{c}_i, y_i, \mathbf{e}_i)\}$, where $\mathbf{c}_i$ is the $i$-th claim to be verified, $y_i \in \{\text{Refute}, \text{Support}\}$ is its ground-truth label, and $\mathbf{e}^{\text{gold}}_i$ is the associated gold evidence. Let $\mathcal{D}_{wrong} = \{i \mid f_{\theta}(\mathbf{c}_i) \neq y_i\}$ denote the subset of instances where a model $f_{\theta}$'s prior-only prediction is incorrect, and reversely, $\mathcal{D}_{correct} = \{i \mid f_{\theta}(\mathbf{c}_i) = y_i\}$ denote the subset where the prior-only prediction is correct. The Correction Rate (\textbf{CR}) measures the model's ability to rectify the initial errors when provided with \emph{correct} external evidence and is computed as the post-retrieval accuracy on $\mathcal{D}_{wrong}$:
\vspace{-0.2em}
    \[
    \small
    \text{CR} = \dfrac{1}{|\mathcal{D}_{wrong}|} \sum_{i \in \mathcal{D}_{wrong}} \mathbb{I} \left[ f_{\theta}(\mathbf{c}_i, \mathbf{e}^{\text{gold}}_i) = y_i \right].
    \vspace{-0.2em}
    \]
The Persistence Rate (\textbf{PR}), however, measures the model's robustness in keeping consistent with the correct prior prediction when exposed to retrieved context with counterfactuals ($\mathbf{e}^{\text{cf}}$ in \S\ref{sec:dataset_construction}). It is computed as the post-retrieval accuracy on $\mathcal{D}_{correct}$:
\vspace{-0.2em}
    \[
    \small
    \text{PR} = \dfrac{1}{|\mathcal{D}_{correct}|} \sum_{i \in \mathcal{D}_{correct}} \mathbb{I} \left[ f_{\theta}(\mathbf{c}_i, \mathbf{e}^{\text{cf}}_i) = y_i \right].
    \vspace{-0.2em}
    \]
\paragraph{Differential Metric (Margin).}
To quantify the benefit of retrieval, we define the \textbf{Margin} as the gap between the model's receptiveness to gold evidence and its susceptibility to discrepant yet counterfactual evidence. %\textcolor{blue}{(i.e., the gap between Knows and Unknows)}. 
Specifically, Margin measures the difference between the rate of desired change (correction) and undesired change (credulousness, i.e., $1 - \text{PR}$):
\vspace{-0.3em}
\[
%\small
\text{Margin} = \text{CR} - (1 - \text{PR}) = \text{CR} + \text{PR} - 1.
\vspace{-0.3em}
\]
A positive Margin indicates that the model is more likely to be corrected by accurate evidence than misled by inaccurate context. Larger Margin values therefore reflect more desirable behavior, combining stronger correction and greater persistence capabilities when facing the discrepancies.

\paragraph{Odds of influence (OI).}
To measure the relative elasticity of a model's parametric confidence, we compare how often it changes versus maintains its prior-based prediction when facing discrepant retrieved context. We define \textbf{OI} as the odds of changing an answer over maintaining it, computed separately under confident (\textsc{Known}) and uncertain (\textsc{Unknown}) conditions for the incorrect (\textsc{Unknows}) and correct (\textsc{Knows}) cases: \textbf{1) \textsc{Unknows} regime} (on $\mathcal{D}_{wrong}$): Since the rate of change is given by CR, the corresponding odds are defined as: 
    $\text{OI}_{\textsc{Known}} = \frac{\text{CR}_{\textsc{Known}}}{1 - \text{CR}_{\textsc{Known}}}$, $\text{OI}_{\textsc{Unknown}} = \frac{\text{CR}_{\textsc{Unknown}}}{1 - \text{CR}_{\textsc{Unknown}}}$. 
    \textbf{2) \textsc{Knows} regime} (on $\mathcal{D}_{\text{correct}}$): Since the rate of change is given by $(1-\text{PR})$, the corresponding odds are: 
    $\text{OI}_{\textsc{Known}} = \frac{1-\text{PR}_{\textsc{Known}}}{\text{PR}_{\textsc{Known}}}$, $\text{OI}_{\textsc{Unknown}} = \frac{1-\text{PR}_{\textsc{Unknown}}}{\text{PR}_{\textsc{Unknown}}}$.
% \vspace{-0.2em}
% \begin{itemize}[leftmargin=*]
%     \item \textbf{\textsc{Unknows} regime} (on $\mathcal{D}_{wrong}$): Since the rate of change is given by CR, the corresponding odds are defined as: 
%     $\text{OI}_{\textsc{Known}} = \frac{\text{CR}_{\textsc{Known}}}{1 - \text{CR}_{\textsc{Known}}}$, $\text{OI}_{\textsc{Unknown}} = \frac{\text{CR}_{\textsc{Unknown}}}{1 - \text{CR}_{\textsc{Unknown}}}$. 
%     \item \textbf{\textsc{Knows} regime} (on $\mathcal{D}_{\text{correct}}$): Since the rate of change is given by $(1-\text{PR})$, the corresponding odds are: 
%     $\text{OI}_{\textsc{Known}} = \frac{1-\text{PR}_{\textsc{Known}}}{\text{PR}_{\textsc{Known}}}$, $\text{OI}_{\textsc{Unknown}} = \frac{1-\text{PR}_{\textsc{Unknown}}}{\text{PR}_{\textsc{Unknown}}}$.
% \end{itemize}
% \vspace{-0.2em}
% \textcolor{red}{Please add a couple of sentences here stating what OI indicates and what value is desired, like the last paragraph of Margin.}
As a tendency to change an answer, a lower OI for \textsc{Knows} regime is preferred to demonstrate persistence (i.e., resisting incorrect context), while a higher OI for \textsc{Unknows} regime is favored to demonstrate receptivity (i.e., accepting correct context to rectify internal errors).

\subsection{Dataset Construction}\label{sec:dataset_construction}
%\textcolor{blue}{We propose \textbf{CAPED} (\underline{C}orrection \underline{A}nd \underline{P}ersistence \underline{E}valuation under prior-context \underline{D}iscrepancy), which is constructed along two primary dimensions: counterfactual and temporal.} 
We construct an evaluation dataset for analyzing LLM's fact-checking behaviors, controllable testing when internal priors and external evidence disagree. The dataset is developed along two primary dimensions (see Figure~\ref{fig:figure2}(upper)), i.e., counterfactual and temporal, to probe fact-checking behavior under diverse prior-context discrepancies. %as shown in Figure~\ref{fig:figure2}(upper).
% The latter is designed to assess model performance when real-world information evolves beyond an LLM’s static training data, a critical scenario in fact-checking where internal knowledge frequently becomes outdated. 
Comparisons with prior datasets are in \S\ref{app: related_works}, with dataset construction and quality control details in \S\ref{app: dataset}.

\paragraph{Dimension 1: Counterfactual.} 
This dimension evaluates model behavior in the presence of erroneous external evidence. We leverage three public fact-checking benchmarks, Quantemp, PolitiFact, and Snopes~\citep{anand2024quantemp,popat-etal-2018-declare}, to construct a consolidated set of 10,866 verifiable claims with ground-truth labels through controlled counterfactual construction and quality control.

\textit{Counterfactual Construction.} Inspired by the controllable discrepancy construction method, like negation injection~\citep{gubelmann2022context}, entity substitution~\citep{wu2024clasheval,longpre-etal-2021-entity,zhou2023context}, and LLM-generated counterfactual~\citep{xie2023adaptive}, 
we formally define counterfactual evidence $\mathbf{e}^{\text{cf}}$ using transformation functions $\mathcal{T}$ applied to the gold evidence $\mathbf{e}^{\text{gold}}$, such that $\mathbf{e}^{\text{cf}}_i = \{\mathcal{T}(\mathbf{e}^{\text{gold}}_i) \mid \mathcal{T} \in \{\mathcal{T}_{\text{entity}}, \mathcal{T}_{\text{semantic}}\}\}$. 
Specifically,
1) \textbf{Counter-entity} ($\mathcal{T}_{\text{entity}}$) replaces key entities in the gold evidence, testing resistance to fine-grained, token-level factual errors;
2) \textbf{Counter-semantic} ($\mathcal{T}_{\text{semantic}}$) 
% Generates counterfactual narratives to test the discernment of broader semantic contradictions.
generates semantically counterfactual evidence that contradicts the original content at a broader semantic level. Each instance in the curated dataset is represented as 
$\{ \mathbf{c}_i, \mathbf{e}^{\text{gold}}_i, \mathbf{e}^{\text{cf\_entity}}_i, \mathbf{e}^{\text{cf\_semantic}}_i\}$,
where $\mathbf{e}^{\text{cf\_entity}}_i$ and $\mathbf{e}^{\text{cf\_semantic}}_i$ correspond to the outputs of $\mathcal{T}_{\text{entity}}$ and $\mathcal{T}_{\text{semantic}}$,  respectively. %\textcolor{red}{The construction method looks thin and unsupported. Can you provide references to back up that this is a standard approach?}

\textit{Quality Control.} 
To ensure a valid evaluation of persistence, we apply a strict refusal filtering procedure, excluding instances where the model abstains (e.g., responses such as ``I cannot answer''), retaining only samples where the model exhibits a definite prior. This creates a genuine tension between internal beliefs and discrepant external context. Counterfactual evidence is generated using GPT-4o under controlled prompting settings. See \S\ref{app: dataset} for more details.
% resulting in a final set of samples evenly split between true and false claims. We exclude instances where the model abstains (e.g., responses such as ``I cannot answer''), retaining only samples where the model exhibits a definite prior. This creates a genuine tension between internal beliefs and discrepant external context. Counterfactual evidence is generated using GPT-4o under controlled prompting settings. See \S\ref{app: dataset} for more details.% and prompts of construction are given in \S\ref{app: dataset}.
% \textcolor{red}{Detailed prompts provided?}. 
\paragraph{Dimension 2: Temporality.} 
This dimension evaluates model behavior when relevant internal knowledge is obsolete or missing, a key requirement for effective RAG-based fact-checking. Temporal PCD is therefore evaluated as a correction rather than a persistence setting: when a claim concerns post-cutoff events, the relevant prior is by construction outdated or unavailable, so persisting in it would not constitute reliable fact-checking.

\textit{Data Collection.} 
We collect valid claims about \textit{new} events by crawling Wikipedia's ``Current events'' portal\footnote{\scriptsize{\url{https://en.wikipedia.org/wiki/Portal:Current_events}}}, focusing on entries from 2024 and 2025. Gold evidence is obtained from the corresponding Wikipedia references. To distinguish between \textit{outdated} and \textit{new} knowledge, we compare each event’s occurrence date with the pre-training cutoff of each evaluated LLM. A claim is considered \textit{new} for a given model if it occurred %strictly 
after the model’s training cutoff, and therefore lies outside its parametric memory. We %further 
retain only claims the model answers \emph{incorrectly} without external evidence (i.e., using its internal priors).
%for which the model consistently produces \emph{incorrect} answers when relying solely on its internal priors (i.e., without external evidence). 
This constraint enforces a valid \textit{acquisition} scenario, requiring the model to resolve the discrepancy by incorporating retrieved context rather than persisting in erroneous priors. The resulting dataset contains 3,995 factually \textit{true} claims. Each instance is structured as 
$\{\mathbf{c}_i, \mathbf{e}^{\text{gold}}_i, T_i \}$, where $\mathbf{e}^{\text{gold}}_i$ denotes the gold evidence for the claim $\mathbf{c}_i$ and $T_i$ is the event timestamp. For fair evaluation, we retain only claims that are considered new for all evaluated LLMs. 
\newcommand{\pct}[1]{#1{\scriptsize\%}}
\newcommand{\sumrow}[1]{{\scriptsize\textcolor{black!75}{#1}}}
\newcommand{\sumpct}[1]{{\scriptsize\textcolor{black!75}{#1\%}}}
% Requires: booktabs, colortbl, arydshln, xcolor

\begin{table*}[t!]
\centering
\scriptsize
\setlength{\tabcolsep}{8pt}
\renewcommand{\arraystretch}{1.05}

\begin{tabular}{ll|*{7}{c}}
\toprule
\multicolumn{2}{l|}{}
& \shortstack{\textbf{Deepseek-v3}}
& \shortstack{\textbf{Gemini-2.5}}
& \shortstack{\textbf{Phi-4}}
& \shortstack{\textbf{Qwen3-32B}}
& \shortstack{\textbf{Mistral-7B}}
& \shortstack{\textbf{GPT-4o-mini}}
& \shortstack{\textbf{Llama3-8B}} \\
\midrule

\multicolumn{9}{c}{\textbf{Dimension 1: Counterfactual Setting}} \\
\midrule

\rowcolor{purple!28}
\multicolumn{2}{l|}{\textbf{Margin} $\uparrow$}
& +0.235 & +0.085 & $-$0.004 & $-$0.115
& $-$0.190 & $-$0.521 & $-$0.661 \\
\midrule
\specialrule{1pt}{1pt}{0pt}

\multicolumn{9}{l}{\quad\textit{\scriptsize
  \textcolor{teal!60!black}{Persistence}
  --- resisting misleading evidence when the prior is correct}} \\[-1pt]

\rowcolor{teal!12}
\textbf{KK} & PR $\uparrow$
& \pct{71.70} & \pct{56.33} & \pct{49.43} & \pct{50.59}
& \pct{71.59} & \pct{40.23} & \pct{29.95} \\
\rowcolor{teal!12}
& $\Delta$ correct drop $\downarrow$
& \pct{17.19} & \pct{21.83} & \pct{11.44}
& \pct{41.79} & \pct{1.55}
& \pct{36.90} & \pct{32.49} \\
\rowcolor{teal!12}
& OI $\downarrow$
& 0.395 & 0.775 & 1.023 & 0.977
& 0.397 & 1.486 & 2.339 \\

% \rowcolor{teal!12}
& Ratio
& \pct{60.75} & \pct{50.00} & \pct{22.62} & \pct{84.57}
& \pct{5.45} & \pct{61.73} & \pct{46.38} \\
\lightsep
\rowcolor{teal!12}
\textbf{UK} & PR $\uparrow$
& \pct{44.05} & \pct{48.50} & \pct{48.84} & \pct{60.82}
& \pct{55.76} & \pct{25.33} & \pct{33.00} \\
\rowcolor{teal!12}
& $\Delta$ correct drop $\downarrow$
& \pct{1.91} & \pct{5.15} & \pct{10.08}
& \pct{1.23} & \pct{4.11}
& \pct{1.33} & \pct{9.96} \\
\rowcolor{teal!12}
& OI $\downarrow$
& 1.270 & 1.062 & 1.048 & 0.644
& 0.793 & 2.948 & 2.030 \\

% \rowcolor{teal!12}
& Ratio
& \pct{3.42} & \pct{10.00} & \pct{19.71} & \pct{3.15}
& \pct{9.29} & \pct{1.78} & \pct{14.87} \\
\lightsep
\rowcolor{gray!36}
\multicolumn{2}{l|}{\sumrow{correct rate (prior) $\uparrow$}}
& \sumpct{64.17} & \sumpct{60.00} & \sumpct{42.33}
& \sumpct{87.72} & \sumpct{14.74}
& \sumpct{63.51} & \sumpct{61.25} \\
\rowcolor{gray!36}
\multicolumn{2}{l|}{\sumrow{correct rate (after intervention) $\uparrow$}}
& \sumpct{45.06} & \sumpct{33.02} & \sumpct{20.81}
& \sumpct{44.70} & \sumpct{9.08}
& \sumpct{25.28} & \sumpct{18.80} \\
% \rowcolor{gray!6}
% \multicolumn{2}{l|}{\sumrow{$\Delta$ correct drop $\downarrow$}}
% & \sumpct{19.11} & \sumpct{26.98} & \sumpct{21.52}
% & \sumpct{43.02} & \sumpct{5.66}
% & \sumpct{38.23} & \sumpct{42.45} \\

\specialrule{1pt}{1pt}{0pt}

\multicolumn{9}{l}{\quad\textit{\scriptsize
  \textcolor{orange!70!black}{Correction}
  --- updating from accurate evidence when the prior is wrong}} \\[-1pt]

\rowcolor{orange!10}
\textbf{KU} & CR $\uparrow$
& \pct{40.75} & \pct{43.65} & \pct{42.76} & \pct{29.75}
& \pct{26.36} & \pct{24.73} & \pct{30.80} \\
\rowcolor{orange!10}
& $\Delta$ wrong reduction $\uparrow$
& \pct{13.10} & \pct{8.37} & \pct{8.26}
& \pct{3.23} & \pct{0.18}
& \pct{8.39} & \pct{8.23} \\
\rowcolor{orange!10}
& OI $\uparrow$
& 0.687 & 0.776 & 0.747 & 0.423
& 0.358 & 0.329 & 0.445 \\
% \rowcolor{orange!10}
& Ratio
& \pct{32.15} & \pct{19.17} & \pct{19.31} & \pct{10.85}
& \pct{0.68} & \pct{33.92} & \pct{26.73} \\
\lightsep

\rowcolor{orange!10}
\textbf{UU} & CR $\uparrow$
& \pct{66.99} & \pct{60.02} & \pct{58.57} & \pct{47.37}
& \pct{27.26} & \pct{57.59} & \pct{40.16} \\
\rowcolor{orange!10}
& $\Delta$ wrong reduction $\uparrow$
& \pct{2.47} & \pct{12.50} & \pct{22.47}
& \pct{0.68} & \pct{23.06}
& \pct{1.48} & \pct{4.83} \\
\rowcolor{orange!10}
& OI $\uparrow$
& 2.030 & 1.502 & 1.414 & 0.900
& 0.375 & 0.345 & 0.672 \\
% \rowcolor{orange!10}
& Ratio
& \pct{3.68} & \pct{20.83} & \pct{38.36} & \pct{1.43}
& \pct{84.58} & \pct{2.57} & \pct{12.02} \\
\lightsep
\rowcolor{gray!36}
\multicolumn{2}{l|}{\sumrow{wrong rate (prior) $\downarrow$}}
& \sumpct{35.83} & \sumpct{40.00} & \sumpct{57.67}
& \sumpct{12.28} & \sumpct{85.26}
& \sumpct{36.49} & \sumpct{38.75} \\
\rowcolor{gray!36}
\multicolumn{2}{l|}{\sumrow{wrong rate (after intervention) $\downarrow$}}
& \sumpct{20.26} & \sumpct{19.13} & \sumpct{26.95}
& \sumpct{8.37} & \sumpct{62.02}
& \sumpct{26.62} & \sumpct{25.69} \\
% \rowcolor{gray!6}
% \multicolumn{2}{l|}{\sumrow{$\Delta$ wrong reduction $\uparrow$}}
% & \sumpct{15.57} & \sumpct{20.87} & \sumpct{30.72}
% & \sumpct{3.91} & \sumpct{23.24}
% & \sumpct{9.87} & \sumpct{13.06} \\

\midrule
\multicolumn{9}{c}{\textbf{Dimension 2: Temporal Setting}} \\
\specialrule{1pt}{1pt}{0pt}

\multicolumn{9}{l}{\quad\textit{\scriptsize
  \textcolor{orange!70!black}{Acquisition}
  --- updating from accurate evidence when the prior is outdated}} \\[-1pt]

\rowcolor{orange!10}
\textbf{KU} & CR $\uparrow$
& \pct{46.58} & \pct{28.65} & \pct{24.39} & \pct{90.90}
& \pct{27.80} & \pct{54.00} & \pct{94.81} \\
\rowcolor{orange!10}
& $\Delta$ wrong reduction $\uparrow$
& \pct{23.71} & \pct{13.89} & \pct{23.13}
& \pct{72.17} & \pct{9.14}
& \pct{46.56} & \pct{35.74} \\
\rowcolor{orange!10}
& OI $\uparrow$
& 0.872 & 0.402 & 0.323 & 9.989
& 0.385 & 1.524 & 9.460 \\
% \rowcolor{orange!10}
& Ratio
& \pct{50.91} & \pct{48.48} & \pct{94.85} & \pct{79.39}
& \pct{32.88} & \pct{86.22} & \pct{37.70} \\
\lightsep

\rowcolor{orange!10}
\textbf{UU} & CR $\uparrow$
& \pct{68.25} & \pct{24.35} & \pct{27.45} & \pct{79.38}
& \pct{40.50} & \pct{68.25} & \pct{98.85} \\
\rowcolor{orange!10}
& $\Delta$ wrong reduction $\uparrow$
& \pct{33.50} & \pct{12.55} & \pct{1.41}
& \pct{16.36} & \pct{27.18}
& \pct{9.40} & \pct{61.58} \\
\rowcolor{orange!10}
& OI $\uparrow$
& 2.150 & 0.322 & 0.378 & 3.850
& 0.681 & 9.230 & 38.37 \\
% \rowcolor{orange!10}

& Ratio
& \pct{49.09} & \pct{51.52} & \pct{5.15} & \pct{20.61}
& \pct{67.12} & \pct{13.78} & \pct{62.30} \\
\lightsep
\rowcolor{gray!36}
\multicolumn{2}{l|}{\sumrow{wrong rate (prior) $\downarrow$}}
& \sumpct{100.00} & \sumpct{100.00} & \sumpct{100.00}
& \sumpct{100.00} & \sumpct{100.00}
& \sumpct{100.00} & \sumpct{100.00} \\
\rowcolor{gray!36}
\multicolumn{2}{l|}{\sumrow{wrong rate (after intervention) $\downarrow$}}
& \sumpct{42.78} & \sumpct{73.57} & \sumpct{75.45}
& \sumpct{11.47} & \sumpct{63.68}
& \sumpct{44.04} & \sumpct{2.67} \\
% \rowcolor{gray!6}
% \multicolumn{2}{l|}{\sumrow{$\Delta$ wrong reduction $\uparrow$}}
% & \sumpct{57.22} & \sumpct{26.43} & \sumpct{24.55}
% & \sumpct{88.53} & \sumpct{36.32}
% & \sumpct{55.96} & \sumpct{97.33} \\

\bottomrule
\end{tabular}

\vspace{-0.5em}
\caption{
Main arbitration profile of LLM verifiers. \capcolorbox{purple!30}{Purple} highlights Margin, the overall
trade-off performance (sum of Known- and Unknown-state margins). \capcolorbox{teal!30}{Teal} rows show persistence
when the prior is correct and counterfactual evidence is given
({KK}/{UK}); PR is persistence rate, OI is odds of influence, and
$\Delta$ correct drop is the correct-rate loss. \capcolorbox{orange!30}{Orange} rows show correction or
acquisition when the prior is wrong or outdated, and gold evidence is given
({KU}/{UU}); CR is correction rate and $\Delta$ wrong reduction is the wrong-rate reduction. \capcolorbox{gray}{Gray} rows give aggregate rates before and after intervention. Ratio reports the percentage of samples falling into each KB condition.
}
\label{tab:main_arbitration_profile}
\vspace{-1.0em}
\end{table*}
\subsection{Evaluated LLMs}
We evaluate seven state-of-the-art LLMs, including closed-source models, such as GPT-4o-mini and Gemini-2.5-flash, and open-source models, such as Mistral-7B (i.e., Mistral-7B-Instruct-v0.2), Llama3-8B, Phi4, Qwen3-32B, and Deepseek-v3, spanning a wide range of scales from 7B to 617B parameters. This diversity enables robust comparison across model architectures and capabilities.

\section{Evaluation Results}\label{main_results}
We organize our analysis into three parts: \textbf{1) cross-model arbitration profiles} 
(Observations 1--3), comparing how different LLMs balance priors and evidence at the model level. \textbf{2) cross-setting arbitration strategies} (Observation 4), exposing how models diverge in handling counterfactual versus temporally new evidence.
% \wg{temporally new? outdated prior?} evidence. 
\textbf{3) modulating factors and model scale} (Observations 5--7), examining how evidence type, label direction, and scaling shape arbitration behavior. All primary results are averaged over ten runs at temperature 0.3; further analysis (e.g., temperature and iteration effects in \S\ref{app:temp}) and details (e.g., prompts) are in \S\ref{our_evaluation_prompts} and \S\ref{more_exp_results}.
%further observation and analysis (the influence of different temperatures and iteration counts (\S\ref{app:temp})), and necessary details (e.g., prompts) are in \S\ref{our_evaluation_prompts} and \S\ref{more_exp_results}.

% We present a comprehensive analysis of experimental results to evaluate how LLMs arbitrate between internal priors and the external evidence under PCD. Primary results are averaged over ten runs with a temperature of 0.3; additional experiments with varying temperatures and iteration counts are detailed in Appendix~\ref{app:temp}. The evaluation set of prompts and further observations is provided in Appendix~\ref{our_evaluation_prompts} and Appendix~\ref {more_exp_results}, respectively.
% All results are averaged over ten runs with a \textcolor{red}{temperature of 0.3. The full set of prompts is provided in Appendix~\ref{more_exp_results}. Additionally, more \textbf{observations} are shown in Appendix~\ref{more_exp_results}}.

% \subsection{Main Results on FactConf}\label{main_results}
 
%\wbshang{[highlight observation and more analysis, such as different LLMs show different knowledge base? different LLMs show different confidence?]}
% \paragraph{The distribution of \textsc{Know} and \textsc{Not-Know} varies across LLMs.}
% \subsection{Cross-Model Arbitration Profiles.}

\paragraph{Observation 1.}~\label{obs:1} \textbf{LLMs exhibit heterogeneous pre-evidence epistemic states.} LLMs display substantial variation in their internal confidence profiles, as manifested by the differing distributions of \textsc{Known} and \textsc{Unknown} categories. This indicates that each model possesses a distinct internal epistemic state. In Table~\ref{tab:main_arbitration_profile}, most evaluated models exhibit high confidence, with more than 80\% of claims falling into \textsc{Known} state. In contrast, Mistral-7B and Phi-4 show significantly higher uncertainty, with around 50\% and 90\% of claims categorized as \textsc{Unknown}, respectively, suggesting limited stability and self-consistency in fact-checking.
Importantly, we analyze the validity of the \textsc{Known}/\textsc{Unknown} classification in \S\ref{app:A3}.

\paragraph{Observation 2.}\label{obs:2} \textbf{High prior-only accuracy does not imply reliable arbitration under PCD.}
Final accuracy alone cannot reveal how a verifier reaches its verdict under PCD.
% \wg{You use source conflict and PCD interchangeably. Is it completely same?}. 
Consider Qwen3-32B and Deepseek-v3: in prior-only % \wg{what is considered clean?}
evaluation, Qwen3-32B substantially outperforms Deepseek-v3 in prior accuracy, positioning it as the stronger verifier under specific benchmarks. Yet this ranking \emph{reverses} once prior-context discrepancy is introduced: Qwen3-32B collapses by 43.02 points, falling below Deepseek-v3 (only a 19.11-point drop). The Margin captures this reversal precisely ($-0.115$ vs.\ $+0.235$), where final-verdict metrics are silent. This reversal is consistent with the variance analysis in %Appendix~\ref{app:A2} 
Table~\ref{tab:combined_variance_stats} (\S\ref{app:A2}): Qwen3-32B's high prior accuracy
is accompanied by high variance ($\sigma=0.2283$ for PR-KK), whereas
Deepseek-v3 combines moderate prior accuracy with low variance
($\sigma=0.0244$). Strong performance in conventional evaluation thus does not guarantee robustness under PCD, underscoring the need for prior-aware evaluation that captures \textbf{\emph{how}} models arbitrate, not just \textbf{\emph{what}} they predict.

\begin{figure}
    \centering
    \includegraphics[width=1\linewidth]{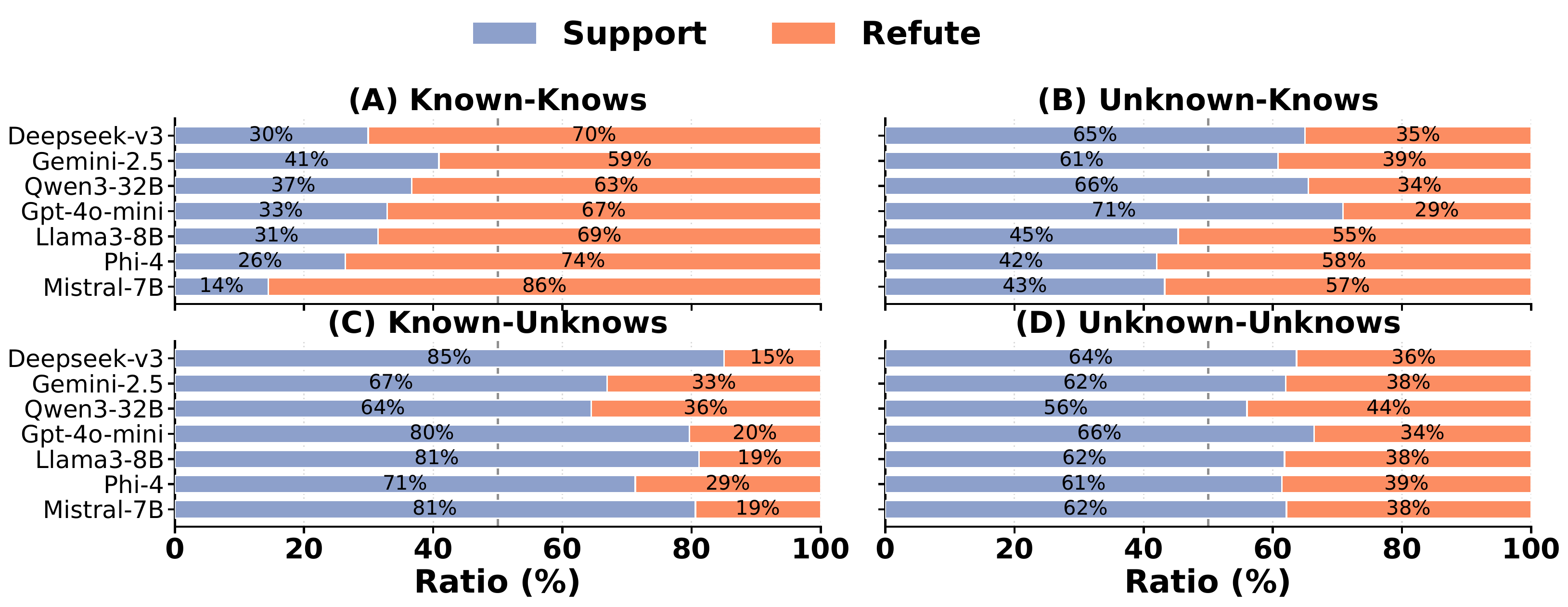}
    \vspace{-1.5em}
    \caption{Support/refute ratios across Known and Unknown prior categories,
    separated by whether the internal prior response is correct or wrong.}
    \label{fig:s_vs_r}
    \vspace{-0.8em}
\end{figure}

\vspace{-0.2em}
\paragraph{Observation 3.}\label{obs:3} \textbf{Model selection is critical for RAG-based fact-checking.} % Not all LLMs are naturally well-suited for handling the discrepancy between internal priors and retrieved context.}
Table~\ref{tab:main_arbitration_profile} illustrates significant variation across LLMs in how they arbitrate internal priors and external evidence, as reflected by their Margin values. Models with positive Margins, such as Deepseek-v3 (+0.235) and Gemini-2.5-flash (+0.085), demonstrate desirable behavior: they resist discrepant context when internally correct, yet remain receptive to corrective evidence when internally incorrect. In contrast, Gpt-4o-mini, Llama3-8B, and Mistral-7B exhibit a more ``stubborn'' profile, frequently favoring erroneous internal priors over accurate external evidence (e.g., Mistral-7B CR: 26.36\%). OI values observed in Table~\ref{tab:main_arbitration_profile} across \textsc{Knows} and \textsc{Unknows} regimes confirm such an observation, where only Deepseek and Gemini demonstrate the favored OI patterns. These results underscore the importance of careful \textbf{\emph{model selection}} for RAG-based fact-checking, considering their distinct PCD handling capabilities.
\vspace{-0.15em}
\paragraph{Observation 4.}\label{obs:4}
% \label{results_on_newfactconf} 
\textbf{Temporal acquisition and counterfactual correction reveal different evidence-uptake patterns.}
At the aggregate level, the OI for acquisition tends to exceed that for correction across models (e.g., GPT-4o-mini: 9.23 vs.\ 0.35; Llama3-8B: 38.37 vs.\ 0.67), suggesting LLMs are more receptive to new knowledge than to revising wrong beliefs. However, this aggregate pattern conceals three distinct strategies. (1) \textbf{Receptive models} (Llama3-8B, Qwen3-32B) drive the trend with large positive gaps (e.g., Llama3-8B's CR: 94.81\% temporal vs.\ 30.80\% counterfactual in KU), readily accepting ``new'' information but resisting evidence that contradicts 
prior knowledge. (2) \textbf{Conservative models} (Gemini-2.5-flash, Phi-4) reverse the pattern, with temporal CR 
\emph{lower} than counterfactual CR (e.g., Phi-4: 24.39\% vs.\ 42.76\% in KU), reflecting a cautious stance toward unfamiliar claims. (3) \textbf{Balanced models} (Deepseek-v3) maintain near-identical correction rates across both settings (46.58\% vs.\ 40.75\%), suggesting genuinely evidence-driven arbitration.
\vspace{-0.25em}
\paragraph{Observation 5.}\label{obs:5} \textbf{LLMs exhibit greater confidence in refuting false claims than in confirming true ones.} %, whereas they exhibit more caution and uncertainty when validating true factual statements.}
% LLMs have a stronger internal prior to identify claims to refute than those to support.}
% LLMs demonstrate greater internal confidence in identifying false facts, whereas they exhibit more caution and uncertainty when validating true factual statements.
Analysis of label distributions in Figure~\ref{fig:s_vs_r} shows that LLMs are more internally confident when identifying false claims than when validating true statements. Among internally correct predictions (i.e., \textsc{Known-Knows}), Refute cases dominate (e.g., 86\% for Mistral-7B and over 59\% for other models), suggesting that pre-trained reliability is more strongly geared toward detecting falsehoods. In contrast, in the \textsc{Unknown} state, correct predictions (i.e., \textsc{Unknown-Knows}) are predominantly Support cases (e.g., 71\% for GPT-4o-mini), indicating heavier reliance on external evidence for verifying affirmative claims where their internal knowledge is insufficient.

\begin{figure}[t!]
    \centering
    \includegraphics[width=1\linewidth]{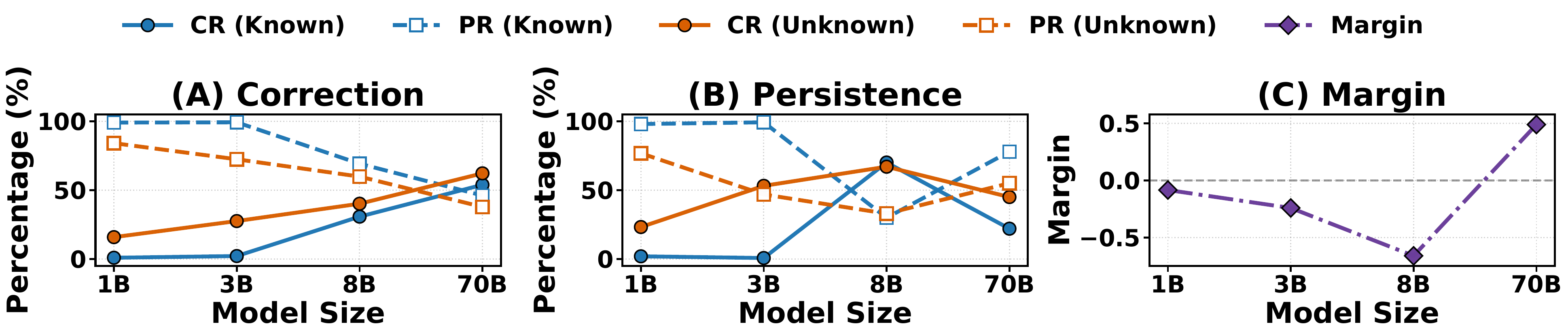}
    \vspace{-1.5em}
    \caption{Model Scaling Evaluation. The Correction rate (CR), persistence rate (PR), and margin results of Llama3 in various sizes (1B, 3B, 8B, 70B)}
    \label{fig:t_and_iteration}
    \vspace{-1.5em}
\end{figure}
\paragraph{Observation 6.} \label{obs:7} \textbf{Larger models exhibit better 
arbitration calibration.}
We analyze how performance evolves with model scale using the Llama3 family. As shown in Figure~\ref{fig:t_and_iteration}, Margin 
scores increase with model size, with Llama3-70B achieving a 
substantial improvement (Margin = 0.50). This trend indicates that 
larger models are more effective at resolving discrepancies. While 
persistence rate decreases from the 3B to the 8B model, it recovers 
at 70B alongside a marked increase in correction rate. The negative 
Margins observed for the 1B and 3B suggest that their persistence 
reflects rigid adherence to incorrect priors rather than robust calibration. 
% Together, this scaling trend suggests that arbitration becomes more genuinely evidence-aware as models grow, rather than reflexively prior-dependent.
This scaling trend suggests that larger models achieve a more
balanced arbitration profile under PCD, improving correction without
entirely sacrificing persistence.

\section{Arbitration Method and Result}\label{method}
Building on these insights, we propose a method to more reliably arbitrate between internal priors and external context, which shows that the diagnostic signal identified by PAVE can be operationalized into a simple test-time intervention. Unlike prior works~\cite{huang2024trust, yu2024truth} that ignore variability across runs, particularly when the model exhibits low internal confidence (i.e., the \textsc{UnKnown} state), 
we assess its decision stability by measuring consistency across multiple parallel generations. Specifically, we compute a Jensen-Shannon divergence (JSD)-based metric to quantify how consistently an LLM favors parametric knowledge or external evidence, enabling more robust source selection. 

\paragraph{Stability-based Arbitration.}
Given a test set $\mathcal{Q} = \{(\mathbf{c}, y)\}$, where $\mathbf{c}$ is a claim and $y \in \{\text{Refute}, \text{Support}\}$ is the ground-truth label, we use an LLM $f_\theta$ to repeatedly sample $K$ times of prediction probabilities for the ``True'' token under external evidence $\mathbf{e}$: 
$\{p_{\text{ext}}^{(i)} = f_\theta(\mathbf{c}, \mathbf{e})\}_{i=1}^K$.
%Given a test set $\mathcal{Q} = \{(\mathbf{c}, \mathbf{y})\}$, where $\mathbf{c}$ is the claim to be checked and the label $\mathbf{y} \in \{\text{Refute}, \text{Support}\}$, we use a large language model $f_\theta$ to sample $N$ output probabilities $\{P_{evi}^i = f_\theta(\mathbf{c}, \mathbf{e})\}_{i=1}^N$ for the ``True'' token given an evidence set $\mathbf{e}$. For each sample, the binary entropy is
For each sample, we define the binary entropy as: 
\begin{equation}\nonumber
\label{eq:entropy}
\small
    H(p_{\text{ext}}^{(i)}) = - p_{\text{ext}}^{(i)} \log_2 p_{\text{ext}}^{(i)} - (1-p_{\text{ext}}^{(i)})\log_2(1-p_{\text{ext}}^{(i)}).
%    \vspace{-0.2em}
\end{equation}
We then compute the JSD across the $K$ samples as follows:
(1) compute the mean probability $\bar{p}_{\text{ext}} = \frac{1}{K} \sum_{i=1}^K p_{\text{ext}}^{(i)}$ and its entropy $H(\bar{p}_{\text{ext}})$;  
(2) compute the expected individual entropy $\mathbb{E}[H(p_{\text{ext}})] = \frac{1}{K} \sum_{i=1}^K H(p_{\text{ext}}^{(i)})$;  
(3) define the stability metric:
% $\mathbf{JSD}_{\text{ext}} = H(\bar{p}_{\text{ext}}) - \mathbb{E}[H(p_{\text{ext}})]$.
%\wei{what ST stands for? Why not just JSD?}\yx{ST stands for stability, is directly using JSD better?}\wei{why `stability', not `consistency'?}\yx{I want to express the stability in the probabilities across the answers, considering that Consistency is already defined as the stability of the answers themselves (true or false)\wei{Too many concepts which appear redundant. There seem no consistency measure defined prior to this point. Just wonder if this measure is just consistency.}.}:
% \vspace{-0.2em}
\begin{equation}\nonumber
\small
\mathbf{JSD}_{\text{ext}} = H(\bar{p}_{\text{ext}}) - \mathbb{E}[H(p_{\text{ext}})],
% \vspace{-0.8em}
\end{equation}
%which corresponds to the JSD for the binary case, quantifying the uncertainty of the model’s outputs given external evidence. 
Lower $\mathbf{JSD}_{\text{ext}}$ indicates higher consistency across runs given external evidence.
Analogously, given $K$ parametric prediction probabilities $\{p_{\text{par}}^{(i)} = f_\theta(\mathbf{c})\}_{i=1}^K$ for the ``True'' token without external evidence, we compute the stability $\mathbf{JSD}_{\text{par}}$ for the parametric case:  
$\mathbf{JSD}_{\text{par}} = H(\bar{p}_{\text{par}}) - \mathbb{E}[H(p_{\text{par}})]$.
%\vspace{-0.2em}
% \begin{equation}\nonumber
% \small
% \mathbf{JSD}_{\text{par}} = H(\bar{p}_{\text{par}}) - \mathbb{E}[H(p_{\text{par}})].
% %\vspace{-0.8em}
% \end{equation}
Finally, we define the difference
$$\Delta= \mathbf{JSD}_{\text{par}} - \mathbf{JSD}_{\text{ext}}$$
and compare it against a threshold $\tau \in [-0.08,-0.06]$, calibrated on a held-out validation set (see \S\ref{our_method_k}), 
% (calibrate range in $[-0.06,-0.08]$) \footnote{We run train-test experiments three times to set hyperparameters $\tau$. Each time renders an optimal $\tau$ that ranges in $[-0.06,-0.08]$, in which $\tau$ is determined using a small set of held-out validation data.}
to determine the reliable source:
% \vspace{-0.5em}
\begin{equation}\nonumber
\small
\hat{y} = \begin{cases}
    f_{\theta}(\mathbf{c}), & \text{if } \Delta \le \tau \\
    f_{\theta}(\mathbf{c},\mathbf{e}), & \text{otherwise.}
\end{cases}
% \vspace{-0.5em}
\end{equation}
% \wei{Any place discussing the setting of $\tau$? Very dangerous if not!}
The full algorithm and implementation details are in Algorithm~\ref{Algorithm} and \S\ref{app:A5}, with a vote-based JSD variant for models without logits in \S\ref{app:voted-based}.
%The full algorithm and other implementation details are provided in Algorithm~\ref{Algorithm} and \S\ref{app:A5}, respectively. An alternative vote-based JSD formulation for models without accessible logits is described in \S\ref{app:voted-based}.

\paragraph{Baselines and Implementation.} We evaluate eight baselines grouped into three categories (details in \S\ref{app:baselines}): (i) \textit{Self-guided methods}~\citep{huang2024trust}: ImplicitSCR prompts the LLM to assess evidence reliability before answering, while ExplicitSCR uses CoT reasoning to compare internal and external context. (ii) \textit{Rule-based methods}~\citep{huang2024trust}: InternalEval and ContextEval switch sources based on self-evaluated response or evidence correctness; InternalConf and ContextConf use token-level confidence thresholds to choose between prior-based and retrieval-based answers~\citep{jiang2023active, huang2024trust}; TPC~\citep{huang2024trust,wu2024clasheval} compares confidence scores using raw or calibrated probabilities. (iii) \textit{Context-based methods}: TACS-LR~\citep{yu2024truth} refines retrieved context using an LLM-based filter before final generation.

\paragraph{Metrics.}
% Given a test set $\mathcal{Q} = {(\mathbf{c}, \mathbf{y})}$, where $\mathbf{c}$ is a claim and $\mathbf{y} \in \{\text{Refute}, \text{Support}\}$, we follow~\cite{huang2024trust, yu2024truth} and evaluate accuracy using three metrics. Let $\mathbf{\hat{y}} = f_{\theta}(\mathbf{c})$ be the parametric answer from the base LLM without evidence
Given a test set $\mathcal{Q} = \{(\mathbf{c}, y)\}$, we follow~\cite{huang2024trust, yu2024truth} and evaluate accuracy under three conditions. Let $\mathbf{e}_c$ and $\mathbf{e}_w$ denote correct and incorrect evidence, respectively, and let the model prediction be $\hat{y} = f_{\theta}(\mathbf{c}, \mathbf{e})$. Using the indicator function $\mathbb{I}(\cdot)$, we define:
% \wei{why do you define this which is not used below?}, with $\mathbf{e}_c$ denoting correct evidence and $\mathbf{e}_w$ denoting incorrect evidence.
i) \textit{Accuracy with correct evidence (CE)}:
$\text{CE} = \frac{1}{|\mathcal{Q}|} \sum_{(\mathbf{c},\, y) \in \mathcal{Q}} \mathbb{I}(f_{\theta}(\mathbf{c}, \mathbf{e}_c) = y)$. 
%\wbshang{[if include the correct result without external evidence, the metric still named correction acc?]}
%\wei{Don't understand this equation: How to explain the probability? why $\mathbb{I}$ inside it?}
ii) \textit{Accuracy with incorrect evidence (IE)}:
$\text{IE} = \frac{1}{|\mathcal{Q}|} \sum_{(\mathbf{c},\, y) \in \mathcal{Q}} \mathbb{I}(f_{\theta}(\mathbf{c}, \mathbf{e}_w) = y)$.
iii) \textit{Overall Accuracy (OA)}:
$\text{OA} = \frac{\text{CE} + \text{IE}}{2}$.
\paragraph{Main Results} Our method achieves strong performance across a range of LLMs, attaining the best or competitive overall accuracy in five out of six evaluations. In Table~\ref{tab:jsd_results}, the largest gains are observed for smaller models such as Mistral-7B, where our approach outperforms the second-best baseline by 6.58 absolute points (57.08\% vs. 50.50\%). Moreover, unlike methods like ContextEval that overly favor either parametric or external signals, our approach maintains a balanced trade-off between correct and incorrect evidence across models, indicating robust arbitration capabilities. For example, GPT-4o-mini achieves strong performance in both settings, with IE 55.38\% and CE 67.14\%, indicating stable arbitration behavior. Results are reported using $K = 5$ runs with temperature set to 1.0. Case studies are provided in \S\ref{sec:case_study}.
\definecolor{DeepPurple}{RGB}{210, 170, 240}
\begin{table}[t!]
\centering
\scriptsize
\setlength{\tabcolsep}{1.5pt}
\renewcommand{\arraystretch}{1.05}
\vspace{-0.3cm}
\begin{tabular}{l|ccc|ccc|ccc}
\toprule
\textbf{Baseline} & \multicolumn{3}{c|}{\textbf{Gpt-4o-mini}} & \multicolumn{3}{c|}{\textbf{Phi-4}} & \multicolumn{3}{c}{\textbf{Mistral-7B}} \\
\midrule
& IE. & CE. & OA. & IE. & CE. & OA. & IE. & CE. & OA. \\
\hline
ImplicitSCR  & 45.48 & 59.88 & 52.68 & 50.10 & 60.80 & 55.45 & 34.56 & 38.83 & 36.70 \\
ExplicitSCR  & 54.12 & 64.35 & 59.24 & 41.81 & 50.84 & 46.33 & 10.33 & 18.77 & 14.55 \\
InternalEval & 51.43 & 57.60 & 54.52 & 47.60 & 48.80 & 48.20 & 6.26  & 3.08  & 4.67  \\
ContextEval  & \underline{55.20} & \underline{67.03} & \underline{61.12} & \textbf{52.50} & \textbf{66.60} & \textbf{59.55} & 27.61 & 10.33 & 18.97 \\
ContextConf  & 47.90 & 64.70 & 56.30 & 52.20 & 52.20 & 52.20 & 15.33 & 10.33 & 12.83 \\
InternalConf & 47.60 & 48.40 & 48.00 & 46.60 & \underline{65.20} & 55.90 & 34.33 & 34.33 & 34.33 \\
TPC          & 46.90 & 61.60 & 54.25 & 51.14 & 51.14 & 51.14 & \underline{43.33} & \underline{57.67} & \underline{50.50} \\
TACS-LR      & 44.60 & 54.00 & 49.30 & 45.20 & 51.80 & 47.15 & 37.20 & 47.80 & 42.50 \\
\rowcolor{DeepPurple!35} Ours (K=5) & \textbf{55.38} & \textbf{67.14} & \textbf{61.26} & \underline{52.40} & 63.60 & \underline{58.00} & \textbf{51.20} & \textbf{62.96} & \textbf{57.08} \\
\midrule
\midrule
\textbf{Baseline} & \multicolumn{3}{c|}{\textbf{Qwen3-32B}} & \multicolumn{3}{c|}{\textbf{Gemini-2.5-flash}} & \multicolumn{3}{c}{\textbf{Llama3-8B}} \\
\midrule
& IE. & CE. & OA. & IE. & CE. & OA. & IE. & CE. & OA. \\
\hline
ImplicitSCR  & 47.77 & 60.08 & 53.93 & 14.90 & 21.85 & 18.38 & 50.40 & 60.00 & 55.20 \\
ExplicitSCR  & 19.27 & 23.55 & 21.41 & \textbf{57.50} & \textbf{66.93} & \textbf{62.22} & 32.60 & 38.60 & 35.60 \\
InternalEval & 52.63 & 59.29 & 55.96 & 49.95 & 55.81 & 52.88 & 51.04 & 58.39 & 54.72 \\
ContextEval  & \textbf{58.49} & \underline{64.25} & \underline{61.47} & \underline{55.61} & 38.73 & 47.17 & \underline{54.92} & \underline{61.17} & \underline{58.05} \\
TACS-LR      & 45.20 & 51.80 & 48.50 & 39.50 & 53.00 & 46.25 & 28.60 & 34.60 & 31.60 \\
\rowcolor{DeepPurple!35} Ours (K=5) & \underline{56.80} & \textbf{69.12} & \textbf{62.97} & 51.12 & \underline{65.01} & \underline{58.07} & \textbf{59.00} & \textbf{65.20} & \textbf{62.10} \\
\bottomrule
\end{tabular}
\vspace{-0.5em}
\caption{Accuracy (\%) under incorrect (IE) and correct evidence (CE), and Overall Accuracy (OA). The results of our method for setting $K=5$.}
\label{tab:jsd_results}
\vspace{-1.5em}
\end{table}
\vspace{-0.5em}
\begin{figure}[t!]
    \centering
    \begin{subfigure}[t]{0.48\textwidth}
        \centering
        \includegraphics[width=\linewidth]{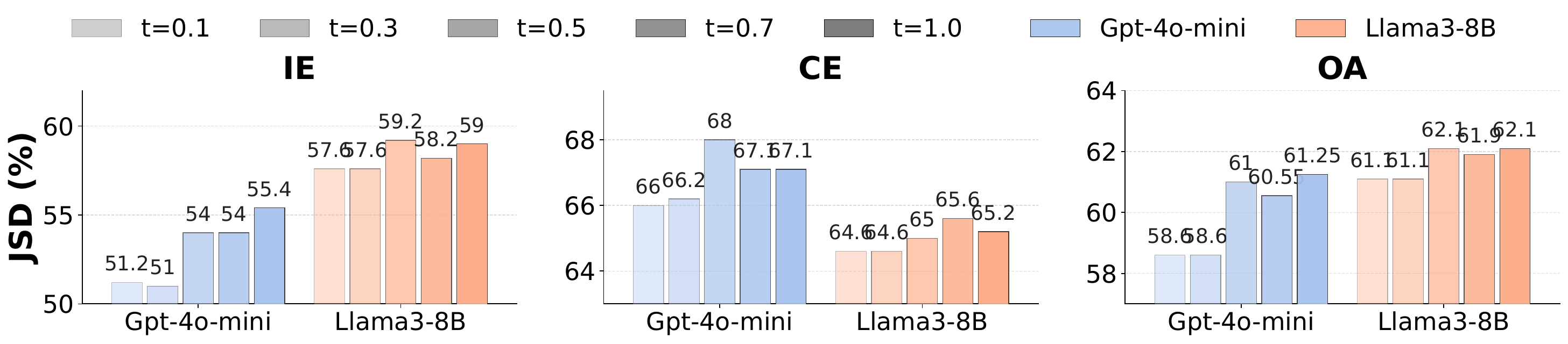}
        \vspace{-1.4em}
        \caption{Different temperature $t \in \{0.1, 0.3, 0.5, 0.7, 1.0\}$.}
        \label{fig:llama3}
    \end{subfigure}
    \hfill
    \begin{subfigure}[t]{0.48\textwidth}
        \centering
        \includegraphics[width=\linewidth]{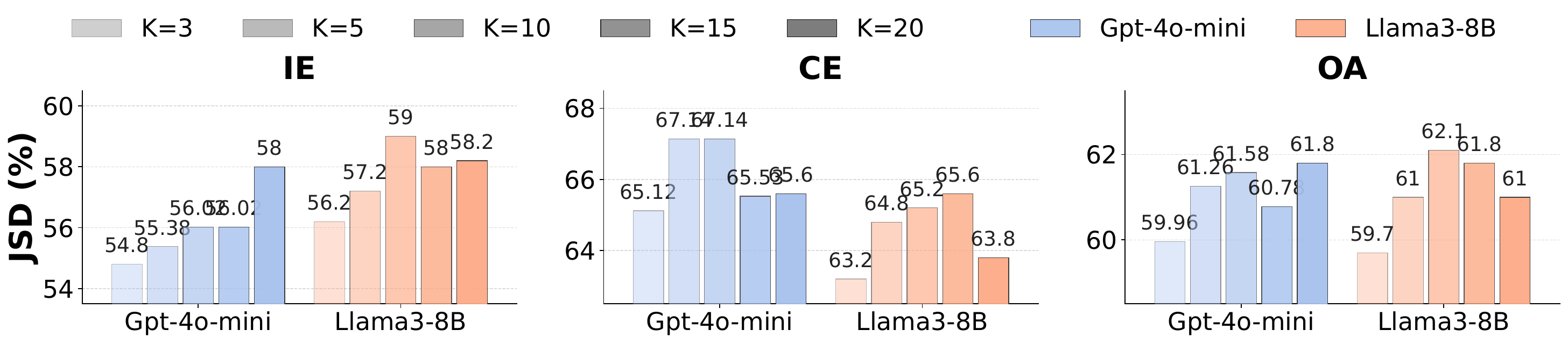}
        \vspace{-1.2em}
        \caption{Number (\#) of runs $K \in \{3, 5, 10, 15, 20\}$.}
        \label{fig:qwen3}
    \end{subfigure}
    \vspace{-0.5em}
    \caption{Comparison of our method across three metrics (IE, CE, and OA) with varying temperature settings (a) and number of runs $K$ (b).}
    \vspace{-1.2em}
    \label{fig:t_and_k}
\end{figure}

% \begin{figure}
%     \centering
%     \includegraphics[width=1\linewidth]{jsd_temperature_and_k_combined.pdf}
%     \caption{Enter Caption}
%     \label{fig:placeholder}
% \end{figure}
\vspace{-0.5em}
\paragraph{Analysis of Efficiency and Temperature.} %As shown in Figure~\ref{fig:t_and_k}, 
Our method demonstrates a favorable efficiency-accuracy trade-off in Figure~\ref{fig:t_and_k}. Increasing the number of runs beyond $K=5$ yields diminishing returns, while even a small budget of $K=3$ achieves competitive performance (e.g., IE = 56.20\%) on Llama3-8B. These results underscore the practicality of the approach for deployment scenarios with limited computational budgets. As for the impact of temperature, results on Llama3-8B and GPT-4o-mini show that increasing temperature from 0.1 to 0.5 leads to noticeable 
performance gains across IE, CE, and OA while no significant or consistent improvement is observed continuously. 
\vspace{-0.5em}
\section{Conclusion}
%This work establishes a rigorous foundation for evaluating the "tug-of-war" between internal priors and external evidence in fact-checking. By introducing the TWINE framework and dataset, we offer deep insights into LLM behaviors across different knowledge boundaries, specifically focusing on correction, persistence, and acquisition. Furthermore, we propose a test-time strategy that effectively adjudicates between parametric and external knowledge, consistently surpassing eight competitive baselines.
This work presents a principled framework for analyzing how LLMs arbitrate between internal priors and retrieved context in fact-checking. We analyze model's behavior under its pre-evidence epistemic states based on Knowledge Boundary, and propose a lightweight test-time arbitration method that outperforms eight strong baselines, improving factual reliability under prior-context discrepancy.

%\wg{Could you check holistically if all the sections of Appendix are properly referred in the main text? Since if some section isn't referred, it wouldn't draw reviewers' attention, as they're not required to read.}\textcolor{purple}{Checked!}

\section*{Limitations}
We acknowledge that different experimental settings may introduce 
variability into evaluation outcomes. We examined three potential 
sources of variation: the choice of output token representation, 
sampling temperature, and the number of independent runs 
(see \S\ref{output variation} and \S\ref{app:temp}). 
While specific metrics fluctuate across configurations, 
the core findings regarding arbitration profiles remain 
relatively stable; detailed sensitivity analyses are 
provided in the appendix.

The primary focus of our dataset is binary fact-checking, 
and our initial results reflect this scope. It should be noted that 
the evaluation of free-form explanations is beyond our current scope. 
Nonetheless, the architecture of our testbed is extensible to 
multi-label tasks; we have included an extensive batch analysis for 
such scenarios in \S\ref{sec:multilabel} to showcase its broader 
applicability.

Our study focuses on the discrepancies between internal priors and 
retrieved context, assuming retrieval has already occurred. 
Consequently, the situation of external conflicts \textit{within} 
the retrieved evidence and agents that autonomously skip the RAG 
process are outside our current scope. We prioritize discrepancy 
scenarios, where one source is correct while the other is not 
(prior vs. context), because they directly impact the finality of 
fact-checking judgments. Situations where both sources are incorrect 
(even if they differ) are excluded, as they do not facilitate 
reaching a correct conclusion. While we recognize the value of 
``both error'' cases for model explainability, a detailed 
investigation into reasoning behavior when all available information 
is erroneous is reserved for future work.

We did not include the cases where LLMs correctly answer 
\textit{new} claims. It is difficult to determine whether these 
correct answers come from valid reasoning (e.g., predicting periodic 
events like the 2026 World Cup) or simply result from hallucinations. 
However, we analyze an example where the LLM correctly predicts 
future events with hallucination in \S\ref{outdated_hallucination}. 
We will leave the analysis of this seemingly ``superior intelligent 
behavior'' for future work.

While our approach entails a marginal increase in overhead compared 
to existing baselines, the empirical results demonstrate that it 
remains highly competitive even at $K=3$, as shown 
in~\S\ref{method} ``Analysis of Efficiency and Temperature''. 
At $K=3$, the method remains practical under batched inference, although it requires repeated stochastic samples under claim-only and evidence-conditioned inputs. Even at this reduced cost, the method retains competitive performance, indicating that our method can strike a practical balance between inference overhead and scoring reliability.

\section*{Ethics Statement}
All evaluations were conducted using open-source datasets and crawled from public websites. We are committed to fostering AI transparency and providing the guidelines of RAG in automated fact-checking systems.
Our data sources are all from objective and neutral facts and do not contain any personal information and offensive comments directed at individuals or particular groups.

Our counterfactual evidence is generated with LLM assistance under controlled prompting. 
While this enables scalable construction of prior-context discrepancy scenarios, it may also introduce generation artifacts, such as stylistic regularities, unnatural wording, or unintended factual changes beyond the targeted intervention.
To mitigate this risk, we apply automatic filtering and human validity checks to remove inconsistent or ill-formed counterfactuals.
We will release the prompts, filtering rules, and constructed data to support reproducibility and facilitate further auditing by the community.
\bibliography{custom}

\newpage
\appendix
% \section*{Appendix}\label{sec:appendix}

\section*{Appendix}
% \paragraph{Organization of the Appendix.}
% This appendix provides supplementary analyses, methodological details, dataset documentation, and reproducibility materials that support the main paper without interrupting its core narrative. 

\paragraph{Appendix Roadmap.}
% To help readers navigate the supplementary material, we provide a brief roadmap
% of the appendix. The appendix is organized into five groups: additional
% empirical analyses, related-work positioning, method details, dataset
% documentation, and reproducibility materials.
This appendix provides supplementary analyses, methodological details, dataset documentation, and reproducibility materials that support the main paper without interrupting its core narrative.
\begin{itemize}[leftmargin=*]
    \setlength{\itemsep}{0.45em}
    \setlength{\topsep}{0.3em}
    \setlength{\parsep}{0em}
    \setlength{\parskip}{0em}

    \item \textbf{Appendix~\ref{more_exp_results}: Further Discussion.}
This section complements the main results with additional empirical analyses
from four perspectives:
\begin{enumerate}[leftmargin=*]
    \setlength{\itemsep}{0.11em}
    \setlength{\topsep}{0.1em}
    \setlength{\parsep}{0em}
    \setlength{\parskip}{0em}

    \item \textbf{More observation.}
    Observation~\ref{obs:6} compares counter-semantic and counter-entity
    conflicts, showing that semantic-level discrepancies are more challenging
    for models to resist.

    \item \textbf{Experimental settings discussion.}
    Appendix~\ref{app:temp} analyzes the effects of temperature and the number
    of independent runs, while Appendix~\ref{app:A3} validates the
    \textsc{Known}/\textsc{Unknown} split through confidence and class-token
    probability distributions.

    \item \textbf{Stability and generality of the evaluation.}
    Appendix~\ref{app:A2} reports variance and standard deviation statistics
    for correction and persistence behaviors across models, and
    Appendix~\ref{sec:multilabel} extends the analysis to multi-label verdict
    prediction on \textsc{PubHealth}.

    \item \textbf{Failure analysis on temporally novel claims.}
    Appendix~\ref{outdated_hallucination} presents hallucination cases where
    models over-confidently extrapolate beyond their verified knowledge
    boundaries, further motivating prior-aware arbitration.

    \item \textbf{Robustness to Output Variation.} Appendix~\ref{output variation} examines whether different output-token choices affect the evaluation results and shows that our findings remain stable across semantically equivalent verbalizers.
\end{enumerate}

    \item \textbf{Appendix~\ref{app: related_works}: Related Works.}
    This section positions our benchmark and our arbitration setting with respect to
    prior work on prior-context discrepancy, knowledge conflicts, and
    uncertainty-based knowledge-boundary estimation.

    \item \textbf{Appendix~\ref{our_method_k}: Further Details and Results of Our Method.}
    % This section gives implementation details and additional analyses of our
    % stability-based arbitration method. 
    Appendix~\ref{app:A5} describes the JSD-based stability measure, while Appendix~\ref{app:voted-based} presents the \textbf{vote-based} variant for models without reliable probability access. The remaining parts provide case studies (App.~\ref{sec:case_study}) and baselines definitions (App.~\ref{app:baselines}).

    \item \textbf{Appendix~\ref{app: dataset}: Dataset Details.}
    This section documents the dataset construction process, including crawled
    emerging-event claims, public fact-checking sources, counterfactual evidence
    generation, and quality-control procedures.

    \item \textbf{Appendix~\ref{our_evaluation_prompts}: Evaluation Prompts and Reproducibility.}
    This section lists the evaluation prompts, counterfactual construction
    prompts, and baseline prompts. The final parts report robustness checks for
    output-token variations and provide the LLM usage statement.
\end{itemize}
\section{Further Discussion} \label{more_exp_results}
% \paragraph{Observation 7.} \textbf{Larger models exhibit greater openness and clearer awareness of internal knowledge limits.}
% % \paragraph{Model Scaling Evaluation.} 
% We analyze how performance evolves with model scale using the Llama3 family. As shown in Figure~\ref{fig:t_and_iteration}, Margin scores increase with model size, with Llama3-70B achieving a substantial improvement (Margin = 0.50). This trend indicates that larger models are more effective at resolving discrepancies. While persistence rate decreases from the 3B to the 8B model, it recovers at 70B alongside a marked increase in correction rate. 
% %We also found an interesting difference: larger models have a higher correction rate. Although Persistence drops between the 3B and 8B models, it increases again at 70B. 
% The negative Margins observed for the 1B and 3B suggest that their persistence reflects rigid adherence to incorrect priors rather than robust accuracy. %"stubbornness" rather than accuracy. 

\begin{figure}[t!]
    \centering
    \includegraphics[width=0.8\linewidth]{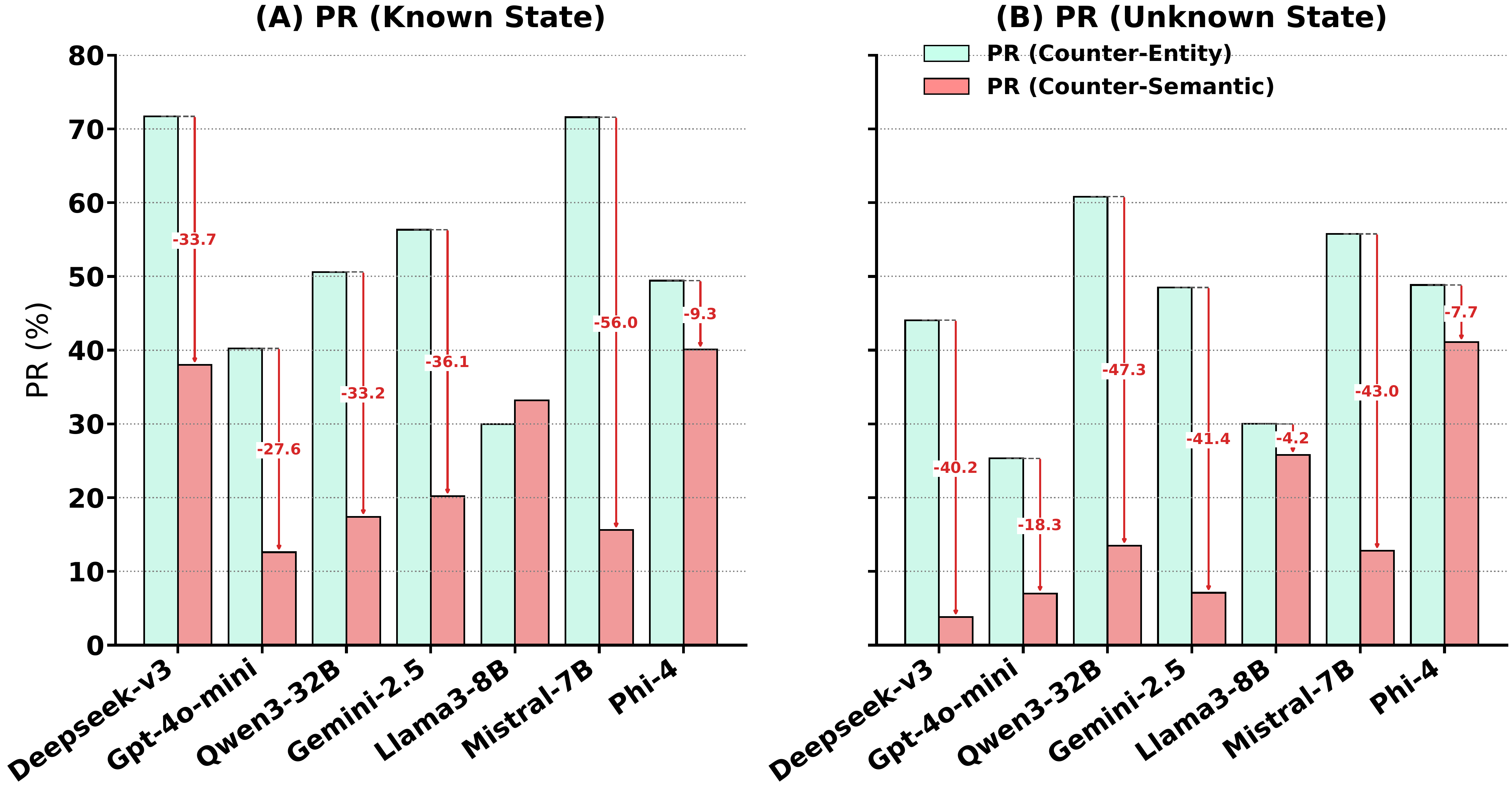}
    % \includegraphics[width=1\linewidth]{imgs/vertical_comparison_plot.pdf}
    % \caption{Persistence of counter-entity vs. counter-semantic.}
    % \vspace{-1.8em}
    \caption{Persistence of counter-entity vs. -semantic.}
    \label{fig:en_vs_se}
    % \vspace{-1.6em}
\end{figure}

\paragraph{Observation 7.}\label{obs:6} \textbf{Counter-semantic counterfactual evidence is much more challenging than counter-entity evidence}. As shown in Figure~\ref{fig:en_vs_se}, most models (with the exception of Llama3-8B) exhibit a notable decrease in persistence rate when exposed to semantically counterfactual evidence compared to entity-level counterfactual evidence (e.g., a 56\% drop for Mistral-7B). This trend, consistent with prior findings by~\citet{xie2023adaptive}, indicates that semantic-level discrepancies pose a greater challenge to current LLMs and are more likely to induce incorrect judgments. Consequently, mitigating such discrepancies may require specialized preprocessing or safeguards, as reliance on internal judgment alone is often insufficient. This confirms the finding in prior work on the difficulty of LLMs autonomously handling deceptive or misleading content~\citep{chen2023can,kamoi2024can,chen2024combating}.

\subsection{Influence of Temperature and \# of Independent Runs} \label{app:temp}

We examine the impact of the \# of independent runs (up to 40) and temperature settings (0.3, 0.5, 1.0) on the consistency of model outputs and the relationship between the influence rate (i.e., correction (CR) and Not persistence (1-PR)) for \textsc{Known} and \textsc{Unknown} cases. The Figure~\ref{fig: t_and_iteration} shows the analysis of Llama3-8B; the proportion of \textsc{Unknown} claims stabilizes after approximately 20 runs (consistently across all three temperature settings), indicating a clear saturation point. 
In the middle and right panels, the influence rate for \textsc{Known} cases decreases steadily with more runs, reflecting growing model confidence, whereas the influence rate of \textsc{Unknown} cases shows only a negligible reduction. This confirms that \textsc{Unknown} effectively captures intrinsic model uncertainty—a state in which further consistency checks do not improve confidence. Influence Rate of \textsc{Unknown} shows limited sensitivity to the number (\#) of runs).

\subsection{Analysis of Known and Unknown Category} \label{app:A3}
\begin{figure}[t!]
    \centering
    \includegraphics[width=0.9\linewidth]{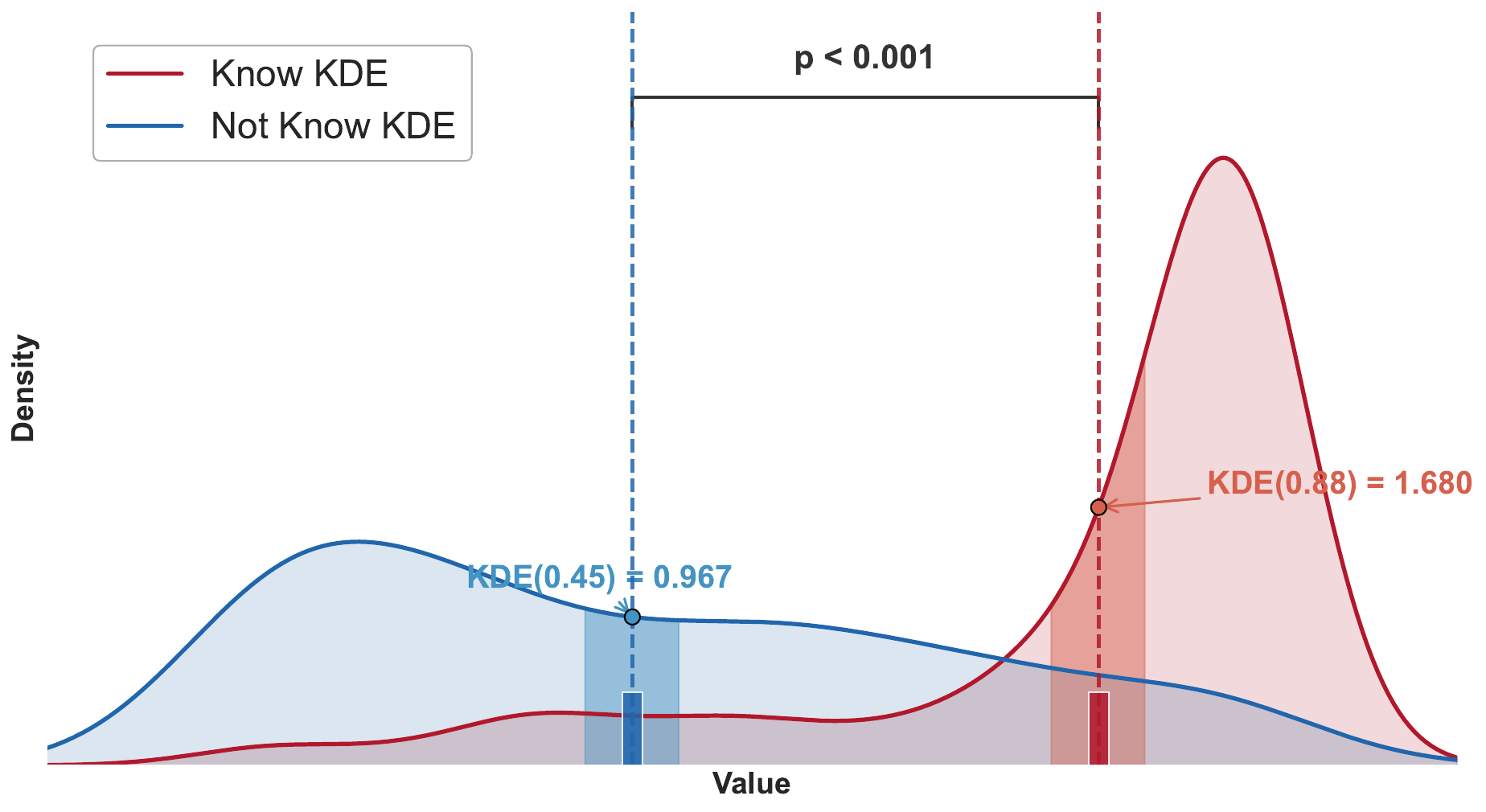}
    \caption{Probability distributions of class tokens (``Support''/``Refute'') predicted by GPT-4o-mini.}
    \label{Fig: know-not-know}
\end{figure}

\begin{figure*}[t!]
    \centering
    \includegraphics[width=1\linewidth]{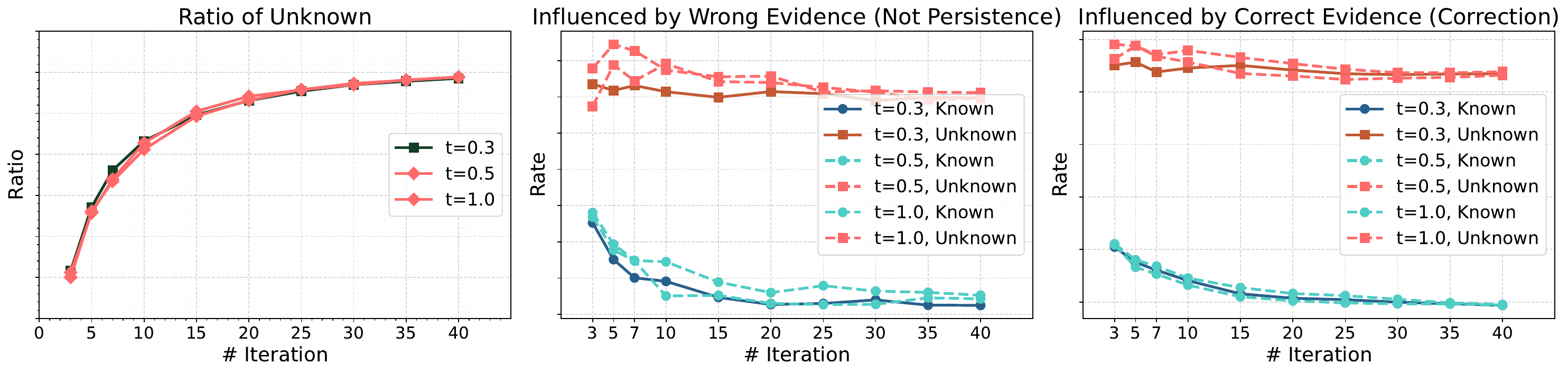}
    \caption{The figure based on the number of independent runs (from 0 to 40) with different temperatures. The proportion of \textsc{Unknown} cases plateaus after reaching a certain number of runs (Left). The influence rate of \textsc{Known} cases decreases more significantly with increasing number of runs when exposed to conflicting evidence, whereas \textsc{Unknown} cases demonstrate a relatively stable influence rate throughout the process (Middle \& Right).}
    % \vspace{-0.3cm}
    \label{fig: t_and_iteration}
    % \vspace{-1em}
\end{figure*}

We experiment by prompting an LLM (e.g., GPT-4o-mini) for binary (Refute/Support) claim verification to measure their internal confidence~\citep{taubenfeld2025confidence, fu2025deep}, which is based on 500 claims, with 5 runs for each one under fixed temperature 1.0. \textsc{Known} cases (red) cluster near $1$, indicating high certainty, while \textsc{Unknown} cases (blue) show broader spread, reflecting uncertainty (Kernel Density Estimation (KDE) smoothing is applied ($p < 0.001$)). Our results show that consistent judgments result in a sharp, low-entropy probability distribution of predicted class tokens, as illustrated in Figure~\ref{Fig: know-not-know}. This indicates the reasonable category of known and Unknown. 
% \begin{figure}[t!]
%     \centering
%     \includegraphics[width=1\linewidth]{imgs/llama_series_metrics.pdf}
%     \caption{Model Scaling Evaluation. The Correction rate (CR), persistence rate (PR), and margin results of Llama3 in various sizes (1B, 3B, 8B, 70B)}
%     \label{fig:t_and_iteration}
%     % \vspace{-0.5em}
% \end{figure}
% Our evaluation with and without external evidence, based on the prompts in Table~\ref{tab:basic_evaluation}, and the entity substitution generation prompts are shown in Table~\ref{tab:entity_replacement_framework}. 
\begin{table}[t!]
\centering

\label{tab:pubhealth_final_debugged}
\small
\renewcommand{\arraystretch}{1.5} % 修复：从 4pt 改为 1.5，保证紧凑且美观
\setlength{\tabcolsep}{3pt}      % 修复：稍微加大间距，避免文字拥挤

\begin{tabular}{l|cc|cc}
\toprule
% \multirow{1}{*}{\textbf{Model}} & \multicolumn{2}{c|}{\textbf{Known}} & \multicolumn{2}{c}{\textbf{Unknown}} \\
% \cmidrule(lr){2-3} \cmidrule(lr){4-5}
% & \textbf{Metric} & \textbf{OI} & \textbf{Metric} & \textbf{OI} \\
% \midrule

% --- 第一部分：Persistence ---
\multicolumn{5}{c}{\cellcolor{gray!10}\textit{Persistence Performance}} \\\hline
\multirow{1}{*}{\textbf{Model}} & \multicolumn{2}{c|}{\textbf{Known-Knows}} & \multicolumn{2}{c}{\textbf{Unknown-Knows}} \\
\midrule
 & \textbf{PR} ($\uparrow$) & \textbf{OI} ($\downarrow$) & \textbf{PR} ($\uparrow$) & \textbf{OI} ($\downarrow$) \\\hline
\textbf{Phi-4}     & 29.50\% & 2.389 & 54.60\% & 0.832 \\
\textbf{Llama3-8B} & 30.71\% & 2.256 & 39.62\% & 1.524 \\

\midrule

% --- 第二部分：Correction ---
\multicolumn{5}{c}{\cellcolor{gray!10}\textit{Correction Performance}} \\\hline
\multirow{1}{*}{\textbf{Model}} & \multicolumn{2}{c|}{\textbf{Known-Unknows}} & \multicolumn{2}{c}{\textbf{Unknown-Unknows}} \\
\midrule
 & \textbf{CR} ($\uparrow$) & \textbf{OI} ($\uparrow$) & \textbf{CR} ($\uparrow$) & \textbf{OI} ($\uparrow$) \\\hline
\textbf{Phi-4}     & 60.53\% & 1.534 & 55.29\% & 1.237 \\
\textbf{Llama3-8B} & 73.59\% & 2.786 & 55.39\% & 1.242 \\

\bottomrule
\end{tabular}

\caption{Consolidated results on the \textsc{PubHealth} dataset. Results are partitioned into Persistence Performance (on correct initial responses) and Correction Performance (on wrong initial responses).}
\label{tab:pubhealth_wrong}
\end{table}

% --- 表 2：Margin 独立表 ---
\renewcommand{\arraystretch}{1.5} % 修复：从 4pt 改为 1.5，保证紧凑且美观
\setlength{\tabcolsep}{3pt} 
\begin{table}[t!]
\centering

\label{tab:pubhealth_margin}
% \small
\begin{tabular}{lclc}
\toprule
\textbf{Model} & \textbf{Margin} $\uparrow$ & \textbf{Model} & \textbf{Margin} $\uparrow$ \\
\midrule
\textbf{Phi-4}     & -0.0008 &\textbf{Llama3-8B} & -0.0069 \\

\bottomrule
\end{tabular}
\caption{The calculated Margin values for each model on the \textsc{PubHealth} dataset, indicating the differentiation capability between internal knowledge and external noise.}
\label{tab:pubhealth_correct}
\end{table}

% \begin{table*}[t!]
% \centering
% \small
% \setlength{\tabcolsep}{6pt} % 稍微减小列间距以适应页面
% \renewcommand{\arraystretch}{1.2}
% \begin{tabular}{c|cccc|cccc}
% \toprule
% \multirow{2}{*}{Model} & \multicolumn{4}{c|}{\textsc{Known}} & \multicolumn{4}{c}{\textsc{Unknown}} \\
% \cline{2-9}
%  & CR Var & CR Std & PR Var & PR Std & CR Var & CR Std & PR Var & PR Std \\
% \hline
% GPT-4o-mini      & 0.0093 & 0.0967 & 0.0034 & 0.0584 & 0.0064 & 0.0799 & 0.0087 & 0.0935 \\
% Deepseek-v3      & 0.0019 & 0.0435 & 0.0006 & 0.0244 & 0.0034 & 0.0585 & 0.0060 & 0.0774 \\
% Qwen3-32B        & 0.0451 & 0.2125 & 0.0521 & 0.2283 & 0.0450 & 0.2120 & 0.0569 & 0.2386 \\
% Gemini-2.5-flash & 0.0129 & 0.1134 & 0.0024 & 0.0495 & 0.0064 & 0.0801 & 0.0049 & 0.0699 \\
% Mistral-7B       & 0.0053 & 0.0729 & 0.0009 & 0.0301 & 0.0035 & 0.0591 & 0.0009 & 0.0293 \\
% Phi-4            & 0.0050 & 0.0704 & 0.0046 & 0.0678 & 0.0033 & 0.0575 & 0.0038 & 0.0616 \\
% Llama3-8B        & 0.0009 & 0.0293 & 0.0009 & 0.0293 & 0.0002 & 0.0135 & 0.0002 & 0.0135 \\
% \bottomrule
% \end{tabular}
% \caption{Variance (Var) and Standard Deviation (Std) for \textbf{Correction Rate (CR)} and \textbf{Persistence Rate (PR)} across \textsc{Known} and \textsc{Unknown} states.\textcolor{blue}{TODO}}
% \label{tab:combined_variance_stats}
% \end{table*}

\begin{table*}[t!]
\centering
\small
\setlength{\tabcolsep}{4pt} % 调整列间距以适应新的布局
\renewcommand{\arraystretch}{1.2}

\begin{tabular}{l|cc|cc||cc|cc}
\toprule
\multirow{3}{*}{\textbf{Model}} & \multicolumn{4}{c||}{\textbf{\small{Persistence Performance (PR)}}} & \multicolumn{4}{c}{\textbf{\small{Correction Performance (CR)}}} \\
\cmidrule(lr){2-5} \cmidrule(lr){6-9}
 & \multicolumn{2}{c|}{\textbf{\textsc{Known-Knows}}} & \multicolumn{2}{c||}{\textbf{\textsc{Unknown-Knows}}} & \multicolumn{2}{c|}{\textbf{\textsc{Known-Unknows}}} & \multicolumn{2}{c}{\textbf{\textsc{Unknown-Unknows}}} \\
\cmidrule(lr){2-3} \cmidrule(lr){4-5} \cmidrule(lr){6-7} \cmidrule(lr){8-9}
 & Var & Std & Var & Std & Var & Std & Var & Std \\
\midrule
% 数据顺序已调整为: [PR-Known] [PR-Unknown] || [CR-Known] [CR-Unknown]
GPT-4o-mini      & 0.0034 & 0.0584 & 0.0087 & 0.0935 & 0.0093 & 0.0967 & 0.0064 & 0.0799 \\
Deepseek-v3      & 0.0006 & 0.0244 & 0.0060 & 0.0774 & 0.0019 & 0.0435 & 0.0034 & 0.0585 \\
Qwen3-32B        & 0.0521 & 0.2283 & 0.0569 & 0.2386 & 0.0451 & 0.2125 & 0.0450 & 0.2120 \\
Gemini-2.5-flash & 0.0024 & 0.0495 & 0.0049 & 0.0699 & 0.0129 & 0.1134 & 0.0064 & 0.0801 \\
Mistral-7B       & 0.0009 & 0.0301 & 0.0009 & 0.0293 & 0.0053 & 0.0729 & 0.0035 & 0.0591 \\
Phi-4            & 0.0046 & 0.0678 & 0.0038 & 0.0616 & 0.0050 & 0.0704 & 0.0033 & 0.0575 \\
Llama3-8B        & 0.0009 & 0.0293 & 0.0002 & 0.0135 & 0.0009 & 0.0293 & 0.0002 & 0.0135 \\
\bottomrule
\end{tabular}
\caption{Variance (Var) and Standard Deviation (Std) for \textbf{Persistence Rate (PR)} and \textbf{Correction Rate (CR)} across Known-Unknown Quadrant.}
\label{tab:combined_variance_stats}
\end{table*}

\subsection{The Variance of Evaluation Results} \label{app:A2}
For the elevation results (Table~\ref{tab:main_arbitration_profile}), we present the variance and standard deviation statistics for the correction rate (CR) and persistence rate (PR) across several language models, shown in Table~\ref{tab:combined_variance_stats}. Overall, the results reveal clear differences in the stability of models' correction and persistence behaviors. Notably, LLama3-8B consistently demonstrates the lowest variance and standard deviation, indicating a highly stable response pattern. In contrast, Qwen3-32b displays the highest variance and standard deviation values, suggesting that its behavior is substantially less consistent and more variable. Some models exhibit similar variance and standard deviation between the correction rate and persistence rate, implying that the volatility in model output affects both metrics equally. In summary, LLama3 and Deepseek-v3 stand out as the most stable models among those evaluated, whereas Qwen3-32b is characterized by pronounced instability.

\begin{table*}[t!]
\centering
\footnotesize

\begin{tabular}{p{0.35\textwidth}|p{0.6\textwidth}}
\hline
\textbf{Claim} & \textbf{LLM Result} \\
\hline
On May 18 2025: 2024–25 Premier League Everton F.C. play their final game at Goodison Park in Walton, Liverpool, England, the club's home ground for the last 133 years. They will transition to Everton Stadium in Vauxhall starting the 2025–26 season. (BBC Sport) & \colorbox{blue!30}{The claim should be supported based on the information available up to} \colorbox{blue!30}{October 2023}. Explanation: Everton FC has been planning to move to a new stadium known as Everton Stadium which is being constructed in the Vauxhall area of Liverpool. The club has announced that they intend to transition to this new stadium, which is expected to be completed in time for the 2025–26 season. As of my last update, the club had been preparing for this significant move, which would mark the end of their long history at Goodison Park, their home since 1892. The specific date mentioned in the claim \colorbox{yellow!30}{(May 18, 2025)} aligns with the conclusion of the 2024–25 Premier League season, \colorbox{yellow!30}{making it plausible that this could be the date} of their final game at Goodison Park. Therefore, the claim is consistent with the club's announcements and plans regarding their stadium transition. \\\hline
% On July 22 2025: 2025 Pakistan floods\nFlash flooding kills three people and leaves 15 others missing in Gilgit-Baltistan, Pakistan. (AP)
% & \colorbox{blue!30}{As of my last knowledge update in october 2023 i cannot provide realtime} \colorbox{blue!30}{information} \colorbox{blue!30}{or verify events that occurred after that date} including specific claims about future events such as the 2025 pakistan floods therefore i cannot definitively support or refute the claim regarding the floods in gilgitbaltistan on july 22 2025\n\nhowever if you are asking for an analysis of the claim based on historical patterns pakistan has experienced significant flooding in the past due to monsoon rains and other climatic factors flash floods in regions like gilgitbaltistan are not uncommon if the claim aligns with historical trends of flooding in that area one might consider it plausible but without specific evidence or reports from that date i cannot confirm its accuracy\n\nin summary i cannot support or refute the claim as it pertains to a future event beyond my last knowledge update.\\
\hline
\end{tabular}

\caption{Data instance with response based on GPT-4o-mini.}
\label{tab:claim_llm_result}
\end{table*}

\subsection{Multi-Label Verdict Prediction}\label{sec:multilabel}

While our current framework is not primarily designed for generating free-form explanations, it is inherently applicable to multi-label classification tasks. To validate its generalizability, we conducted supplementary experiments by collecting the \textsc{PubHealth} dataset~\citep{kotonya-toni-2020-explainable-automated}, involving 3,000 claims. This task requires models to classify claims into multiple categories, such as ``True,'' ``False,'' and ``Unproven.'' The detailed results are summarized in Table~\ref{tab:pubhealth_correct} and Table~\ref{tab:pubhealth_wrong}.

We observe that when applied to this multi-label scenario, the performance Margin for Phi-4 remains negative, whereas for Llama3-8B, it remains positive. These findings are highly consistent with their respective behaviors in the binary-label tasks presented in the main manuscript, further confirming the robustness of our stability-based arbitration strategy across different classification granularities.

\subsection{Hallucination Analysis}~\label{outdated_hallucination}
We analyze the results with an explanation based on GPT-4o-mini in \textit{novel} claims from our benchmarks and results shown in Table~\ref{tab:claim_llm_result}. The LLM response explicitly states a knowledge cutoff date of October 2023 but proceeds to evaluate a claim involving a specific future date: May 18, 2025. This date corresponds to the purported final Everton F.C. match at Goodison Park. While the model correctly references Everton's publicly known stadium relocation plans and the typical timing of the Premier League season’s conclusion, its concrete acceptance of May 18, 2025 as a plausible final match date exceeds verified knowledge boundaries. Specifically, since official fixture lists and exact match dates for the 2024–25 Premier League season would likely only be confirmed and publicly released in mid-2025, the model’s certainty about the precise date is unsupported by direct evidence available before the cutoff. The response frames the date as “plausible” but ultimately judges the claim as “should be supported,” which risks conflating reasonable inference with factual confirmation.

This constitutes a form of \textit{hallucination}, where the model goes beyond its verified knowledge to make a detailed prediction without authoritative backing. Although not fabricated outright, it represents an overconfident extrapolation that could mislead users into treating the claim as a confirmed fact rather than an informed speculation.

\begin{table}[t!]
\centering

\label{tab:kappa_summary}
% \small
\begin{tabular}{lc}
\toprule
\textbf{Indicator} & \textbf{Value} \\
\midrule
Fleiss' Kappa $\uparrow$ & 0.8856 \\
Observed Agreement ($P_o$) & 0.9830 \\
Expected Agreement ($P_e$) & 0.8517 \\
Number of Samples & 465 \\
Number of Raters (Prompts) & 10 \\
\bottomrule
\end{tabular}
\caption{Inter-agreement analysis (Fleiss' Kappa)}
\label{fleiss}

\end{table}
\begin{table*}[t!]
\centering
\footnotesize
\label{tab:kappa_matrix}
\resizebox{\textwidth}{!}{
\begin{tabular}{lcccccccccc}
\toprule
 & \textbf{Accurant} & \textbf{accurant} & \textbf{Correct} & \textbf{correct} & \textbf{True} & \textbf{true} & \textbf{Support} & \textbf{support} & \textbf{Right} & \textbf{right} \\
\midrule
\textbf{Accurant} & - & 0.8458 & 0.9173 & 1.0000 & 0.8458 & 0.8458 & 1.0000 & 0.8458 & 0.7832 & 0.7832 \\
\textbf{accurant} & 0.8458 & - & 0.9275 & 0.8458 & 1.0000 & 1.0000 & 0.8458 & 0.8632 & 0.9353 & 0.9353 \\
\textbf{Correct}  & 0.9173 & 0.9275 & - & 0.9173 & 0.9275 & 0.9275 & 0.9173 & 0.9275 & 0.8634 & 0.8634 \\
\textbf{correct}  & 1.0000 & 0.8458 & 0.9173 & - & 0.8458 & 0.8458 & 1.0000 & 0.8458 & 0.7832 & 0.7832 \\
\textbf{True}     & 0.8458 & 1.0000 & 0.9275 & 0.8458 & - & 1.0000 & 0.8458 & 0.8632 & 0.9353 & 0.9353 \\
\textbf{true}     & 0.8458 & 1.0000 & 0.9275 & 0.8458 & 1.0000 & - & 0.8458 & 0.8632 & 0.9353 & 0.9353 \\
\textbf{Support}  & 1.0000 & 0.8458 & 0.9173 & 1.0000 & 0.8458 & 0.8458 & - & 0.8458 & 0.7832 & 0.7832 \\
\textbf{support}  & 0.8458 & 0.8632 & 0.9275 & 0.8458 & 0.8632 & 0.8632 & 0.8458 & - & 0.8058 & 0.8058 \\
\textbf{Right}    & 0.7832 & 0.9353 & 0.8634 & 0.7832 & 0.9353 & 0.9353 & 0.7832 & 0.8058 & - & 1.0000 \\
\textbf{right}    & 0.7832 & 0.9353 & 0.8634 & 0.7832 & 0.9353 & 0.9353 & 0.7832 & 0.8058 & 1.0000 & - \\
\bottomrule
\end{tabular}
}
\caption{Pairwise Kappa agreement matrix across different output tokens}
\label{kappa}
\end{table*}

\subsection{Robustness to Output Token Variations}~\label{output variation}
Previous studies have noted that model performance can exhibit significant variance depending on the specific input prompts and the chosen output tokens during constrained decoding~\cite{mahaut2024factual}. As shown in Table~\ref{fleiss} and \ref{kappa}, to rigorously evaluate Mistral-7B's sensitivity to output formats, we conducted a robustness experiment on a 500-sample subset. We designed 10 distinct prompts by varying the output constraints across five semantic pairs—\textit{accurate/inaccurate}, \textit{true/false}, \textit{correct/incorrect}, \textit{support/refute}, and \textit{right/wrong}—including their capitalized variants. With the temperature set to 0.5, we measured inter-prompt consistency using Fleiss' Kappa. As shown in Tables~\ref{kappa} and \ref{fleiss}, the results indicate a very high level of agreement. The overall Fleiss' Kappa across all 10 prompts reached \textbf{0.8856}, with all pairwise Kappa scores exceeding 0.78 and many surpassing 0.95. This demonstrates that the model's fact-checking performance is robust and not a byproduct of specific output token selection.

% \subsection{Multi-label}
% 请确保导言区加载了以下宏包：
% \usepackage{booktabs}
% \usepackage{multirow}
% \usepackage{caption}

% \begin{figure}
%     \centering
%     \includegraphics[width=1\linewidth]{iclr2026/imgs/all_metrics_with_t_comparison_phi4.png}
%     \caption{Comparison of iteration number $N$ under different temperature settings (Phi4) when temperature = 0.3.}
%     \label{fig:placeholder}
% \end{figure}

\section{Related Works}\label{app: related_works}

\paragraph{Datasets of PCD.} To the best of our knowledge, there is currently a lack of datasets specifically designed for fact-checking tasks within the context of Prior-Context Discrepancy (PCD). To address this gap, we introduce our benchmarks, a large-scale dataset engineered to evaluate how models navigate discrepancies between their internal memory and external information. While related research has explored model behavior in other overlapping scenarios, our benchmarks provides a more comprehensive and multi-dimensional analysis. Compared to representative datasets such as ClashEval~\citep{wu2024clasheval}, DynamicQA~\citep{marjanovic-etal-2024-dynamicqa} and \citep{xie2023adaptive}, our benchmarks distinguishes itself through its extensive coverage of complex reasoning scenarios. Specifically, it evaluates model performance on retrieved new events where internal parametric knowledge is outdated, and systematically measures the impact of varying counterfactual context, including both counter-entity (CE) and counter-semantic (CS) counterfactual. Furthermore, different from~\citep{su2024conflictbank, xie2023adaptive,ming2024faitheval,marjanovic-etal-2024-dynamicqa}, our benchmarks is unique in its ability to verify knowledge persistence—evaluating whether a model can consistently utilize accurate external knowledge to rectify its erroneous internal priors. Crucially, all evaluation dimensions within our benchmarks are validated across the four distinct internal epistemological states of the model: Known-Knows, Known-UnKnows, Unknown-Knows, and Unknown-UnKnows. This allows for a granular assessment when faced with varying degrees of epistemic certainty.

\paragraph{Estimation Knowledge Boundary via Uncertainty.} In the context of identifying knowledge boundaries~\cite{li2024knowledge,hu2023uncertainty}, Uncertainty Estimation (UE) serves to quantify a model’s lack of confidence in its predictions for a given input~\citep{li2024knowledge,hu2023uncertainty}. High uncertainty typically signifies that the input-related knowledge lies outside the model's certain boundaries. Specifically, the epistemic uncertainty plays an important role in UE, which refers to model-specific uncertainty that quantifies a lack of model knowledge, a concept directly linked to the Parametric Knowledge Boundary~\citep{li2024knowledge}.

Current solutions to quantify epistemic uncertainty are broadly classified into model-side and data-side approaches. Model-side quantification involves the perturbation of model parameters, configurations, or internal states. Data-side quantification encompasses two primary modes: Input-side clarification and perturbation, adjusting the input prompts to detect fluctuations in confidence~\citep{gao-etal-2024-spuq,ling-etal-2024-uncertainty}. Output-side variation estimation, quantifying uncertainty based on the diversity or inconsistency of generated outputs~\citep{yadkori2024believe,aichberger2024many}.

Our framework specifically utilizes output-side variation estimation~\citep{li2025knowledge}, a consistency-based method where the inconsistency among multiple sampled predictions for a single input is viewed as the measure of uncertainty. In the specialized domain of fact-checking, consistency-based methods are uniquely sufficient for identifying knowledge boundaries. Fact-checking tasks target objective, verifiable truths where, ideally, only one correct verdict exists. If a model possesses the necessary parametric knowledge, its reasoning and final verdict should remain stable across multiple samplings. It indicates that the model is ``guessing'' due to a knowledge deficit rather than retrieving a stable fact, thereby making output variation a robust proxy for mapping the model's knowledge boundary without requiring access to internal model logits.

\section{Further Details and Results of Our Method}~\label{our_method_k}

\begin{algorithm*}[t!]
\caption{Evidence Stability via Vote Entropy (vote-based)}
\begin{algorithmic}[1]
\Require Claim $\mathbf{c}$, Retrieved Evidence $\mathbf{e}$, an LLM $f_\theta$, iteration times $K$, Threshold $\tau$
\Ensure Decision: Provide the answer $\hat{y}$ based on voting consistency
\State Initialize counters: $k_{\mathrm{ext}} \gets 0$, $k_{\mathrm{par}} \gets 0$ \Comment{Count of 'True' labels}
\State Initialize label sets: $Y_{\mathrm{ext}} \gets \emptyset$, $Y_{\mathrm{par}} \gets \emptyset$
\For{$i=1$ to $K$}
    \State $y_{\mathrm{ext}}^{(i)} \gets f_\theta(\mathbf{c}, \mathbf{e})$ \Comment{Generate hard label with evidence}
    \State $y_{\mathrm{par}}^{(i)} \gets f_\theta(\mathbf{c})$ \Comment{Generate hard label without evidence}
    \If{$y_{\mathrm{ext}}^{(i)} \in \text{Positive Categories}$} $k_{\mathrm{ext}} \gets k_{\mathrm{ext}} + 1$ \EndIf
    \If{$y_{\mathrm{par}}^{(i)} \in \text{Positive Categories}$} $k_{\mathrm{par}} \gets k_{\mathrm{par}} + 1$ \EndIf
\EndFor
\State Compute empirical probabilities: $\hat{p}_{\mathrm{ext}} \gets \frac{k_{\mathrm{ext}}}{K}$,\quad $\hat{p}_{\mathrm{par}} \gets \frac{k_{\mathrm{par}}}{K}$
\State Compute vote entropy (stability):
\State $V_{\mathrm{ext}} \gets H_{\mathrm{binary}}(\hat{p}_{\mathrm{ext}}) = -\hat{p}_{\mathrm{ext}} \log_2(\hat{p}_{\mathrm{ext}}) - (1 - \hat{p}_{\mathrm{ext}}) \log_2(1 - \hat{p}_{\mathrm{ext}})$
\State $V_{\mathrm{par}} \gets H_{\mathrm{binary}}(\hat{p}_{\mathrm{par}}) = -\hat{p}_{\mathrm{par}} \log_2(\hat{p}_{\mathrm{par}}) - (1 - \hat{p}_{\mathrm{par}}) \log_2(1 - \hat{p}_{\mathrm{par}})$
\State $\Delta = V_{\mathrm{par}} - V_{\mathrm{ext}}$ \Comment{Positive $\Delta$ means evidence increases consistency}
\If{$\Delta \le \tau$}
    \State \Return $\hat{y} = \text{Majority}(Y_{\mathrm{par}})$ \Comment{Evidence introduces too much disagreement}
\Else
    \State \Return $\hat{y} = \text{Majority}(Y_{\mathrm{ext}})$ \Comment{Evidence leads to a more stable consensus}
\EndIf
\end{algorithmic}
\label{Algorithm:VoteBased}
\end{algorithm*}
\begin{figure*}[t!]
    \centering
    \includegraphics[width=1\linewidth]{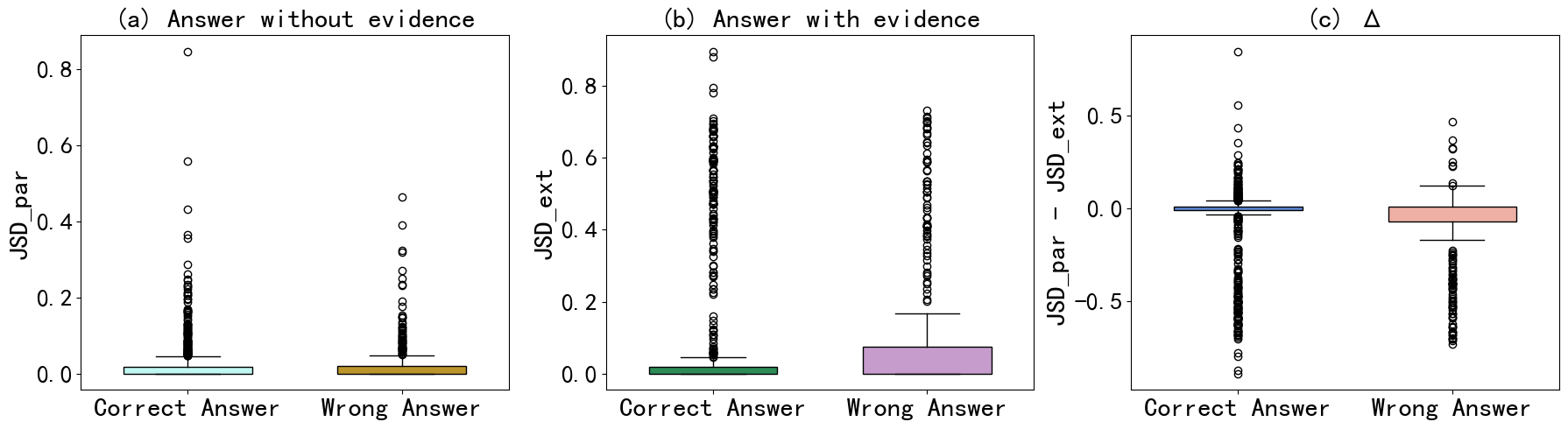}
    \caption{The comparison of our defined JSD with and without evidence.}
    \label{fig:jsd}
\end{figure*}

\subsection{The Implementation Details of JSD}\label{app:A5}
Our method aims to assess the internal consistency of a model’s answers by leveraging the answer probability associated with each response and the algorithm shown in Algorithm~\ref{Algorithm}.

As shown in Table~\ref{tab:jsd_results}, our method achieves competitive performance across six diverse language models, demonstrating remarkable generalization capability. It attains the best overall accuracy on GPT-4o-mini, Mistral-7B, Qwen3-32B, and Llama3-8B, while securing second place on Phi-4. The most striking advantage emerges on smaller-scale models: on Mistral-7B, our approach outperforms the second-best method by 6.58 absolute percentage points, revealing its particular efficacy in resource-constrained environments. This scalability advantage is crucial for practical deployments where computational efficiency is paramount. Furthermore, our method maintains balanced performance between correct evidence (CE) and incorrect evidence (IE) handling across most models, indicating robust reasoning capabilities rather than over-reliance on evidence quality. The sole exception occurs with Gemini-2.5, where ExplicitSCR performs best, suggesting model-specific optimization opportunities. We discuss the influence of different iteration times as follows. 

We run train-test experiments three times to set hyperparameters $\tau$. Each time renders an optimal $\tau$ that ranges in $[-0.08,-0.06]$, in which $\tau$ is determined using a small set of held-out validation data.

As shown in Figure~\ref{fig:jsd}, we compare the Jensen–Shannon Divergence (JSD) $\nonumber
\mathbf{JSD}_{\text{}} = H(\bar{p}_{\text{}}) - \mathbb{E}[H(p_{\text{}})]$ with correct evidence under two settings: answering with the claim only (Figure~\ref{fig:jsd}(a)) and answering with both the claim and external evidence (Figure~\ref{fig:jsd}(b)), using GPT-4o-mini under correct and incorrect response conditions. In the claim-only scenario (Figure~\ref{fig:jsd}(a)), the JSD values are similar regardless of whether the model’s answer is correct or incorrect. In contrast, when external evidence is provided (Figure~\ref{fig:jsd}(b)), the JSD is noticeably lower when the model’s answer is correct than when it is incorrect. This pattern suggests that GPT-4o-mini may possess a certain capability to discern evidence quality: while external evidence may still influence the model, it also introduces greater internal uncertainty, as reflected by the elevated JSD in incorrect responses.

% \begin{figure}
%     \centering
%     \includegraphics[width=1\linewidth]{iclr2026/imgs/our_method_compare.jpeg}
%     \caption{Comparison of our method in different \# of runs $K$ based on GPT-4o-mini and Llama3-8B.}
%     \label{fig: our_more_results}
% \end{figure}
% \paragraph{Analysis of different \# of runs $K$.} We evaluate the performance of our method across different runs ($K = {5, 10, 15, 20}$) using Llama3-8B and GPT-4o-mini as benchmark models. As shown in Figure~\ref{fig: our_more_results}, our approach achieves competitive performance when $K=5$. With further increases in the number of runs, only marginal improvements are observed, indicating that performance does not exhibit consistent enhancement or degradation with additional iterations.

% \subsection{The implementation details of our method.} 
\begin{algorithm*}[t!]
% \footnotesize
\caption{Evidence Stability via Jensen-Shannon Divergence (standard)}
\begin{algorithmic}[1]
\Require Claim $\mathbf{c}$, Retrieved Evidence $\mathbf{e}$, an LLM $f_\theta$, iteration times $K$, Threshold $\tau$
\Ensure Decision: Provide the answer $\hat{y}$ by using or not using external evidence $\mathbf{e}$
\State Initialize arrays $\{p_{\mathrm{ext}}^{(i)}\}_{i=1}^K$, $\{p_{\mathrm{par}}^{(i)}\}_{i=1}^K$
\For{$i=1$ to $K$}
 \State $p_{\mathrm{ext}}^{(i)} \gets f_\theta(\mathbf{c}, \mathbf{e})$ \Comment{Sample with evidence}
 \State $p_{\mathrm{par}}^{(i)} \gets f_\theta(\mathbf{c})$ \Comment{Sample without evidence}
\EndFor
\State Compute mean probabilities: $\bar{p}_{\mathrm{ext}} \gets \frac{1}{K}\sum_{i=1}^K p_{\mathrm{ext}}^{(i)}$,\quad $\bar{p}_{\mathrm{par}} \gets \frac{1}{K}\sum_{i=1}^K p_{\mathrm{par}}^{(i)}$
\State Compute evidence stability: $\mathbf{JSD}_{\mathrm{ext}} \gets H(\bar{p}_{\mathrm{ext}}) - \frac{1}{K}\sum_{i=1}^K H(p_{\mathrm{ext}}^{(i)})$ \Comment{Using the binary entropy function}
\State Compute parametric stability: $\mathbf{JSD}_{\mathrm{par}} \gets H(\bar{p}_{\mathrm{par}}) - \frac{1}{K}\sum_{i=1}^K H(p_{\mathrm{par}}^{(i)})$ \Comment{Using the binary entropy function}
\State $\Delta= \mathbf{JSD}_{\text{par}} - \mathbf{JSD}_{\text{ext}} $
\If{$\Delta \le \tau$}
 \State Do not use evidence, \Return $\hat{y} = f_\theta(\mathbf{c})$ \Comment{External evidence introduces variability}
\Else
 \State Use evidence, \Return $\hat{y} = f_\theta(\mathbf{c}, \mathbf{e})$ 
\EndIf
\end{algorithmic}
\label{Algorithm}
\end{algorithm*}
\subsection{The Vote-based JSD}\label{app:voted-based}
 While Our method aims to assess the internal consistency of a model’s answers by leveraging the answer probability associated with each response, not all models provide direct access to reliable answer probabilities. To address this limitation, we introduce a simple variant of our approach that estimates stability using the results of multiple consistency checks for models that can not access the probability directly.

Specifically, we approximate the external stability measure \(\mathbf{JSD}_{\text{}} = H(\bar{p}_{\text{}}) - \mathbb{E}[H(p_{\text{}})]\) using vote entropy—a frequency-based entropy computed from hard labels. Formally, given a set of \(K\) labels (in $K$ runs), let $k_{\text{true}} \in [0, K]$  denote the number of labels that belong to any of the true categories. Then, the empirical probability is estimated as \(\hat{p} = k_{\text{true}} / K\), and the vote entropy is computed as the binary entropy of \(\hat{p}\): $H_{\text{binary}}(\hat{p}) = -\hat{p} \log_2(\hat{p}) - (1 - \hat{p}) \log_2(1 - \hat{p})$ This variant allows us to robustly evaluate stability even when explicit probability outputs are unavailable. The algorithm is shown in Algorithm~\ref{Algorithm:VoteBased}.

% \begin{figure}
%     \centering
%     \includegraphics[width=1\linewidth]{iclr2026/imgs/2MI_comparison_boxplot_big.png}
%     \caption{Enter Caption \yx{to be revised MI}}
%     \label{fig:jsd}
% \end{figure}

\begin{table*}[t!]

\centering
\scriptsize
\begin{tabular}{>{}l |p{10.6cm}}
\toprule
\hline
% Field & Value \\
% \midrule
Claim ($\mathbf{c}$) & High school students arrested on campus are twice as likely not to graduate and four times less likely to graduate if they’ve appeared in court. \\
External Evidence $\mathbf{e}$ & \begin{minipage}[t]{10.6cm} high school dropouts are three and one-half times more likely than high school graduates to be arrested, and more than eight times as likely to be incarcerated. 12 pages, seventy-five middle school students who were under court have been involved in the judicial system, making sure they ``complete college''. High school dropouts are three and one-half times more likely than high school graduates to be arrested, and more than eight times as likely to be incarcerated. \small{\textcolor{red}{$\times$}} \end{minipage} \\
Initial Answer $f_{\theta}(\hat{y}\mid \mathbf{c}_i)$ & ``support'', ``support'', ``support'', ``support'', ``support'', ``support'', ``support'', ``support'', ``support'', ``support'' \\ 
Initial Answer Probabilities  & 0.9965, 0.9950, 0.9971, 0.9858, 0.9950, 0.9965, 0.9965, 0.9965, 0.9971, 0.9958 \\
Answer with evidence $f_{\theta}(\hat{y}\mid \mathbf{c}_i, \mathbf{e}_i)$ & ``refute'', ``refute'', ``refute'', ``refute'', ``refute'', ``refute'', ``refute'', ``refute'', ``refute'', ``refute'' \\
Evidence Probabilities & 0.9399, 0.9241, 0.4115, 0.4620, 0.9399, 0.9399, 0.8808, 0.4620, 0.9399, 0.4046 \\
$\mathbf{JSD}_{\text{ext}}$ (claim only) & 0.0011 \\
$\mathbf{JSD}_{\text{par}}$ (claim with conflicting evidence) & 0.2239 \\
$\Delta= \mathbf{JSD}_{\text{ext}} - \mathbf{JSD}_{\text{par}}$  & $\approx -0.2229$ \textcolor{textred}{(Choose answer with claim only)} \\
\midrule
Claim ($\mathbf{c}$) & Trade agreements like NAFTA and permanent normal trade relations with China, which forced American workers to compete against people making pennies an hour, has resulted in the loss of 160,000 jobs here in Michigan.'' \\
External Evidence $\mathbf{e}$ & \begin{minipage}[t]{9.5cm}trade agreements like NAFTA and permanent normal trade relations with China, which forced American workers to compete against people making pennies an hour, have resulted in the loss of 160,000 jobs here in Michigan. The steel industry setbacks account for just a fraction of the job losses in Michigan's manufacturing sector, which now employs 55,100 fewer workers than it did when Trump took office in January 2017, U.S. Labor Department data shows. The state's automotive industry accounted for 35\% of the manufacturing job losses, according to the st. Louis fed. Trade agreements like NAFTA and permanent normal trade relations with China, which forced American workers to compete against people making pennies an hour, have resulted in the loss of 160,000 jobs here in Michigan. \small{\textcolor{green}{\checkmark}} \end{minipage} \\
Initial Answer (10 times) & ``refute'', ``refute'', ``refute'', ``refute'', ``support'', ``refute'', ``support'', ``support'', ``refute'', ``support'' \\
Initial Answer Probabilities $f_{\theta}(\hat{y}\mid \mathbf{c})$ & 0.9193, 0.6770, 0.6183, 0.6171, 0.3707, 0.6770, 0.8121, 0.3707, 0.6199, 0.8121 \\
Answer with Evidence $f_{\theta}(\hat{y}\mid \mathbf{c}_i, \mathbf{e})$ & ``support'', ``refute'', ``support'', ``support'', ``support'', ``support'', ``support'', ``support'', ``support'', ``support'' \\
Evidence Probabilities & 0.8175, 0.1480, 0.8519, 0.9526, 0.8175, 0.9526, 0.8519, 0.9046, 0.8519, 0.9046 \\
$\mathbf{JSD}_{\text{ext}}$ (claim only) & 0.1438 \\
$\mathbf{JSD}_{\text{par}}$ (claim with conflicting evidence) & 0.0165 \\
$\Delta= \mathbf{JSD}_{\text{ext}} - \mathbf{JSD}_{\text{par}}$  & $\approx 0.1273$ \textcolor{textred}{(Choose answer with evidence)}\\
\midrule
Claim ($\mathbf{c}$) & Says under President Barack Obama, the debt increased by 23 percent, which was less than under any president going back to Ronald Reagan. \\
External Evidence ($\mathbf{e}$)& \begin{minipage}[t]{10.6cm}27 Feb 2023 under President Barack Obama, the national debt grew from \$10.63 to \$19.96 trillion, a 87.8\% increase. Under President Donald Trump, the ... Oct 7, 2022, Bush nearly matched the amount of debt accumulated under Reagan, but ... Bush, democratic president barack obama added another \$8.34 trillion ... May 15, 2023, Ronald Reagan and George W. Bush. President Ronald Reagan increased the U.S. debt by around 1.86 trillion U.S. dollars, or 186.36 percent. this ...\small{\textcolor{green}{\checkmark}} \end{minipage} \\
Initial Answer (10 times) & ``refute'', ``support'', ``refute'', ``support'', ``refute'', ``support'', ``refute'', ``support'', ``refute'', ``support'' \\
Initial Answer Probabilities $f_{\theta}(\hat{y}\mid \mathbf{c}_i)$ & 0.3760, 0.8485, 0.3760, 0.6764, 0.3760, 0.6183, 0.3760, 0.8485, 0.4360, 0.7230 \\
Answer with Evidence $f_{\theta}(\hat{y}\mid \mathbf{c}_i, \mathbf{e}_i)$ & ``refute'', ``refute'', ``refute'', ``refute'', ``refute'', ``refute'', ``refute'', ``refute'', ``refute'', ``refute'' \\
Evidence Probabilities & 0.9999973, 0.9999973, 0.9999973, 0.9999971, 0.9999973, 0.9999973, 0.9999981, 0.9999973, 0.9999971, 0.9999971 \\
$\mathbf{JSD}_{\text{ext}}$ (claim only) & 0.0316 \\
$\mathbf{JSD}_{\text{par}}$ (claim with conflicting evidence) & 2.5305e-08 \\
$\Delta= \mathbf{JSD}_{\text{ext}} - \mathbf{JSD}_{\text{par}}$  & $\approx 0.0316$ \textcolor{textred}{(Choose answer with evidence)}\\
\hline
\bottomrule
\end{tabular}
\caption{Here are examples of our method for selecting answers, with and without supporting evidence, demonstrated across 10 independent runs. Our approach can successfully identify the correct initial answer even when the external evidence is inaccurate (indicated by \textcolor{red}{$\times$}). Additionally, it selects the answer backed by evidence when that evidence is correct (indicated by \textcolor{green}{\checkmark})}.
\label{tab:case_study}
\end{table*}

\subsection{Case Study}\label{sec:case_study}
% \section{Case Study: Empirical Validation of Stability-Based Evidence Selection}
% \label{sec:case_study}

To empirically validate the effectiveness of our stability-based evidence selection framework, we present a comprehensive case study analyzing three distinct factual verification scenarios. These cases demonstrate how our method dynamically arbitrates between parametric knowledge and external evidence based on output stability metrics. As shown in Table~\ref{tab:case_study}.

The first case involves a claim regarding high school arrest statistics and graduation rates, where the provided external evidence contains internal contradictions (marked with \textcolor{red}{$\times$}). Our method computes a negative $\Delta$ value ($-0.2229$), correctly identifying that the evidence introduces substantial variability ($\mathbf{JSD}_{\mathrm{par}} = 0.2239$) compared to the highly stable parametric responses ($\mathbf{JSD}_{\mathrm{ext}} = 0.0011$). Consequently, the system rejects the conflicting evidence and maintains the original ``support'' classification based solely on parametric knowledge.

In the second case examining trade agreement impacts on employment, the external evidence provides consistent factual support (marked with \textcolor{green}{$\checkmark$)}. Here, the parametric responses show significant instability ($\mathbf{JSD}_{\mathrm{ext}} = 0.1438$) with mixed ``refute''/``support'' predictions, while evidence-augmented queries yield stable consensus ($\mathbf{JSD}_{\mathrm{par}} = 0.0165$). The positive $\Delta$ ($0.1273$) correctly triggers evidence utilization, shifting the majority prediction from ``refute'' to ``support.''

The third case, concerning national debt growth under different presidential administrations, further illustrates the method's precision. Despite initial parametric uncertainty (mixed responses with $\mathbf{JSD}_{\mathrm{ext}} = 0.0316$), the external evidence produces near-perfect stability ($\mathbf{JSD}_{\mathrm{par}} \approx 2.53 \times 10^{-8}$). The positive $\Delta$ ($0.0316$) leads to evidence adoption, resulting in unanimous ``refute'' decisions that correctly contradict the original claim.

These cases collectively demonstrate three critical capabilities: (1)~the ability to detect and reject unreliable evidence that introduces instability; (2)~the capacity to identify and utilize high-quality evidence that enhances prediction consistency; and (3)~the robustness to make appropriate evidence selection decisions across diverse factual domains without requiring explicit credibility assessments of the evidence content. The stability metric $\Delta$ serves as an effective proxy for evidence quality, enabling reliable arbitration between internal knowledge and external information sources.

\subsection{Baseline Methods}\label{app:baselines}

% We compare several state-of-the-art methods for context reliability assessment. Some baselines need to assess the probability of LLMs, such as ContextConf, TPC, InternalConf, therefore, such method we just examined on GPT-4o-mini, Phi-4 and Mistral-7B, which we can obtain the probability. As shwon in Table~\ref{tab:jsd_results}.

Following~\citet{huang2024trust}, we evaluate eight methods for evidence reliability assessment under six prominent LLMs, grouped into three categories: \textbf{Self-guided methods}: \textbf{1)} ImplicitSCR: The LLM is prompted with a claim and potentially incorrect evidence, then judges the reliability of the evidence before producing a final response. \textbf{2)} ExplicitSCR: The LLM generates answers using internal knowledge and external context, performs explicit confidence reasoning via CoT and answer comparison, and then balances both sources to make a final decision. \textbf{Rule-based methods}: \textbf{3)} InternalEval: The LLM evaluates its own initial parametric response; if deemed correct, it is kept, otherwise a new answer is generated using external evidence. \textbf{4)} ContextEval: The LLM evaluates external evidence; if correct, the evidence is used, otherwise it responds only with the claim. \textbf{5)} InternalConf~\citep{jiang2023active}: The LLM chooses between internal and evidence-based answers based on token-level probabilities, using a predefined threshold. \textbf{6)} ContextConf: Similar to InternalConf but applied to evidence-first responses, i.e., the selection is based on prompting with evidence first. \textbf{7)} TPC: Compares confidence scores between internal and context-based answers, using raw probabilities or calibrated percentiles. \textbf{Context-based methods}: 
\textbf{8)} TACS-LR: Filters unreliable context using LLM-based removal, then uses the refined context to generate the final answer~\citet{yu2024truth,huang2024trust}. Consequently, these methods are evaluated only on GPT-4o-mini, Phi-4, and Mistral-7B, from which such probability data can be obtained. The comprehensive results are presented in Table~\ref{tab:jsd_results}.

\begin{itemize}[leftmargin=*]
\item \textbf{Implicit Self-Guided Confidence Reasoning (ImplicitSCR):} The model is explicitly instructed that the provided context may contain inaccuracies and must exercise independent judgment to evaluate its reliability before formulating a response. To prioritize the activation of internal knowledge, the context is deliberately positioned after the question. Some works~\citep{huang2024trust} demonstrate that this sequential structure increases the model’s reliance on prior knowledge, thereby reducing its vulnerability to misleading contextual information. The model subsequently engages in implicit confidence reasoning during its decision process and returns only the final answer. Prompts shown in Table~\ref{tab:ImplicitSCR and ExplicitSCR}.

\item  \textbf{Explicit Self-Guided Confidence Reasoning (ExplicitSCR):} In this approach, the model initially generates distinct responses derived from its internal knowledge and the provided context, using two separate prompts. It subsequently engages in verbalized confidence estimation through a chain-of-thought reasoning process. This begins with an evaluation of the confidence in its internally generated answer, including reflection on the factual basis underlying that response. The model then assesses the reliability of the external context by comparing it against the facts supporting its internal knowledge. The final answer is selected through a deliberative reasoning process that integrates both internal certainty and contextual credibility. Prompts shown in Table~\ref{tab:ImplicitSCR and ExplicitSCR}.
    
\item  \textbf{Internal Evaluation (InternalEval):} The LLM assesses the correctness of its internal answer through self-evaluation using the prompt provided in Table~\ref{tab:InternalEval and ContextEval}. If the self-evaluation confirms the answer is correct, the internal answer is retained; otherwise, the context-based answer is selected.
    
\item  \textbf{Context Evaluation (ContextEval):} The model evaluates the relevance and accuracy of the provided context in relation to the question (see Table~\ref{tab:InternalEval and ContextEval} for the prompt specification). If the context is judged to be correct and reliable, the context-based answer is adopted; otherwise, the model defaults to its internal knowledge-based response.
    
\item  \textbf{Internal Confidence Thresholding (InternalConf):} The model selects its internal knowledge-based answer if its confidence exceeds a predefined threshold; otherwise, it defaults to the context-based answer. The threshold can be set to a fixed value (e.g., 0.5) or calibrated on a held-out dataset when available. Confidence estimates may be derived from sequence probabilities. Following ActiveRAG~\citep{jiang-etal-2023-active}, \citet{huang2024trust} simplifies to this internal confidence mechanism (InternalConf) to employ answer-level probability, yielding modest but consistent empirical improvements. 
    
\item \textbf{Context Confidence Thresholding (ContextConf):} The model evaluates the reliability of the external context. If confidence in the context exceeds a predefined threshold, the context-based answer is selected; otherwise, the model defaults to its internal knowledge-based response. The threshold can be determined using the same methodology as in Internal Confidence estimation. Confidence scoring may be derived from the sequence probability of the answer given the context.

\item \textbf{(Calibrated) Token Probability Correction (TPC):} Following~\citep {wu2024clasheval}, compare the confidence scores—specifically, the mean token probabilities—of the model’s internal answer and the context-based answer, selecting the one with the higher value as the final answer. This approach is termed token probability correction.
    
\item \textbf{Truth-Aware Context Selection (TACS-LR)} Following~\citep {yu2024truth}, who provide Truth-Aware Context Selection~\citep{yu2024truth}, a context evaluation method, which employs a classifier to filter out incorrect content at a granular sentence or token level. Unlike rule-based approaches that simply choose between internal and context-based answers, this method reintegrates the filtered context into the LLM to regenerate the final answer. ~\citet{huang2024trust} adopts this to an alternative strategy: the model explicitly removes untruthful sentences (TACS-LR; see Table~\ref{tab:talr_framework} for the prompt), and the refined context is used to produce the final answer. 
\end{itemize}

\section{The Details of Datasets}\label{app: dataset}
To ensure the quality of automatically constructed conflicting evidence, we established a multi-stage quality control process to verify its plausibility and reliability. The generated evidence underwent systematic filtering and validation under multiple constraints: during generation, the LLM was instructed to ensure that substituted entities shared the same type as the original and that at least ten key entities could be extracted from the context to guarantee semantic richness. After automatic construction, we conducted manual sampling reviews by three human annotators. This rigorous workflow ensures the high quality and challenging nature of the final dataset.

% ~\label{details_newfactconf}
% \begin{table*}[htbp!]
% \centering
% \tiny
% \label{tab:data_example}
% \caption{Example of our collected events with evidence, which is new to Phi-4, Deepseek-v3, etc.}
% \begin{tabular}{p{3.8cm}| p{5.5cm} | p{1cm} | p{2cm}}
% \toprule
% \textbf{Claim} & \textbf{Evidence} & \textbf{Time} & \textbf{Outdated for LLMs}\\
% \midrule
% % \cellcolor{lightblue}
% July 31,2025. The military junta of Myanmar formally ends the country's four-year-long state of emergency and declares a December 2025 election for the country's new head of government and legislative members.  & ... Myanmar's military government plans to hold a general election for elected seats in the Amyotha Hluttaw and the Pyithu Hluttaw of the Assembly of the Union, currently dissolved, on a date in December 2025 to be determined. The planned election would be the first after the 2021 military coup d'état ... & July 31, \newline 2025 &  GPT-4o-mini, \newline LLama3-8B \newline Phi-4 \newline Deepseek-v3,\newline Gemini-2.5-flash \newline Mistral-7B \\
% \bottomrule
% \end{tabular}
% \label{Table: examples_db2}
% \end{table*}
% \begin{wrapfigure}{r}{0.45\textwidth}
    \begin{figure}
    \centering
    \includegraphics[width=1\linewidth]{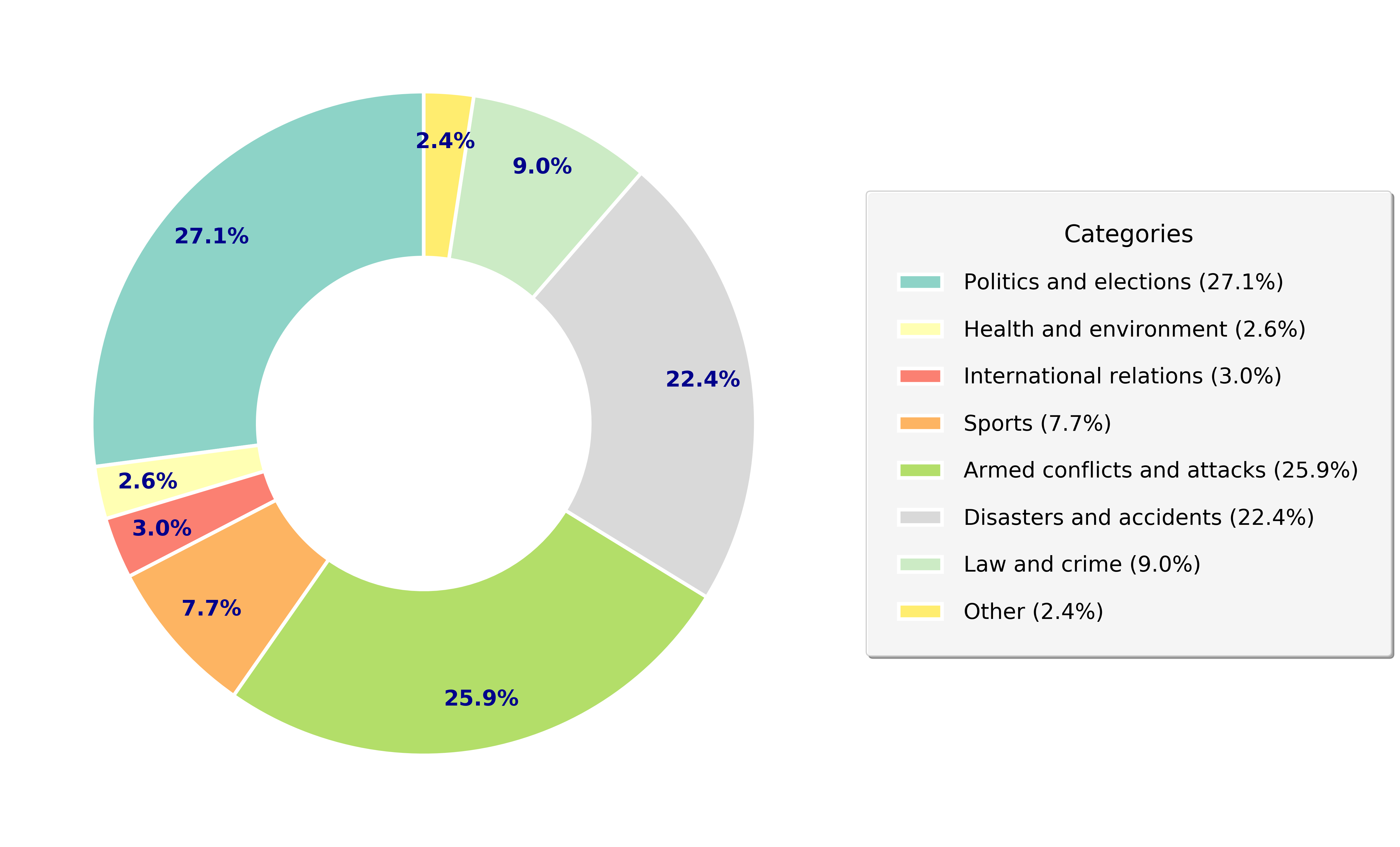}
    \caption{The categories of our crawled data.}
    \label{fig:newfactvonf_details}
\end{figure}
% \end{wrapfigure}
\paragraph{Crawled data.} Our crawled dataset, which contains 3,995 true claims, is used to analyze LLM performance on unseen emerging events; the examples are shown in Table~\ref{App: examples_db2_table}. Concretely, we collected \textit{new} events as claims by crawling the Wikipedia current events portal in different months in 2024 and 2025 (e.g., \url{https://en.wikipedia.org/wiki/Portal:Current_events/June_2025}). If one LLM, such as Llama3-8B, which is published in April 2024, then the parametric knowledge is deemed to be unseen for the events that occurred in June 2025. So we annotated the events which crawled from the web page June 2025 is \textit{new} to Llama3-8B. 
% Then, as shown in Table~\ref{Table: examples_db_fact_conf}, we systematically created conflicting evidence for every claim using the entity-substitution protocol described in previous experiments. 
The category distributions are displayed in Figure~\ref{fig:newfactvonf_details}, and all the categories are sourced from Wikipedia.

\paragraph{The data collection.} We collect three public datasets:
\begin{itemize}
    \item QuanTemp~\citep{anand2024quantemp} is a diverse, multi-domain dataset exclusively focused on numerical claims. It encompasses comparative, statistical, interval, and temporal aspects, and includes detailed metadata along with a supporting evidence collection. 
    \item Snopes: We use the dataset from \cite{popat2017truth}, which consists of rumors from the general fact-checking website Snopes (www.snopes.com). Each instance includes an editor-assigned credibility label (true or false), alongside associated reporting articles and their sources.
    \item PolitiFact: \citet{popat-etal-2018-declare} extracted a dataset from PolitiFact (www.politifact.com), which focuses on claims made by U.S. political figures. Our dataset includes all articles published prior to December 2017, each containing the claim, the speaker, and an official credibility rating.
\end{itemize}

\paragraph{Counterfactual data quality control.}
We provide additional examples of the constructed counterfactual evidence in Table~\ref{Table: examples_db_fact_conf}. 
To improve the plausibility and reliability of the generated counterfactual evidence, we apply a multi-stage quality control procedure. 
During generation, we impose type and complexity constraints: for entity-level counterfactuals, the substituted entity must have the same semantic type as the original entity; for semantic-level counterfactuals, the generated evidence must remain relevant to the original claim while introducing a factual conflict. 
We also require that at least five key entities can be extracted from the context, ensuring that the evidence contains sufficient factual structure for controlled intervention. 
And the detailed prompts of dataset construction in~\ref{tab:entity_replacement_framework} and \ref{tab:counterfactual_evidence_prompt}.

\paragraph{Human validity audit.}
Because our counterfactual evidence is generated with LLM assistance, we conduct a human validity audit to verify that the constructed evidence induces the intended prior-context discrepancy, rather than introducing unrelated content or obvious generation artifacts. 
The audit focuses on whether each counterfactual instance carries the targeted factual intervention.

We randomly sample $M{=}100$ instances from the final counterfactual set, stratified by data source (QuanTemp / PolitiFact / Snopes), counterfactual type (entity-level / semantic-level), and gold label (\textsc{Support} / \textsc{Refute}). 
Each instance is independently annotated by three annotators along four binary criteria: 
(i) {claim relevance}, whether the evidence still concerns the same claim or event; 
(ii) {intervention consistency}, whether it introduces the intended factual conflict; 
(iii) {label consistency}, whether it provides the intended misleading verification signal for the original claim; and 
(iv) {artifact-free plausibility}, whether it is fluent and free of obvious generation artifacts. 
An instance is counted as valid if the majority of annotators pass the first three criteria.

As shown in Table~\ref{tab:cf_human_audit}, the criterion-level pass rates range from 88.0\% for label consistency to 95.0\% for claim relevance, with substantial inter-annotator agreement (Fleiss' $\kappa{=}0.76$). 
Entity-level counterfactuals score higher than semantic-level ones, as entity substitutions are more tightly constrained. 
The lower label consistency rate reflects cases where the intended factual intervention is present but does not unambiguously flip the verification signal, which is a known challenge in counterfactual data construction~\cite{zmigrod-etal-2019-counterfactual}. 
Overall, the audit suggests that the benchmark largely satisfies its intervention design, while semantic-level counterfactuals remain more challenging to construct reliably.

\begin{table}[t]
\centering
\small
\begin{tabular}{lccc}
\toprule
Criterion & Entity-CF & Semantic-CF & Overall \\
\midrule
Claim relevance           & 96.0 & 94.0 & 95.0 \\
Intervention consistency  & 94.0 & 88.0 & 91.0 \\
Label consistency         & 90.0 & 86.0 & 88.0 \\
Artifact-free plausibility& 94.0 & 90.0 & 92.0 \\
\midrule
Fleiss' $\kappa$          & 0.78 & 0.73 & 0.76 \\
\bottomrule
\end{tabular}
\caption{Human audit of counterfactual evidence quality ($M{=}100$, 3 annotators per instance).
We report majority-vote pass rates (\%) and inter-annotator agreement (Fleiss' $\kappa$).
An instance is retained as valid if the majority vote passes the first three criteria.}
\label{tab:cf_human_audit}
\end{table}

\section{Evaluation Prompts}~\label{our_evaluation_prompts}
Overall, we provide the prompts of basic evaluation Table~\ref{tab:basic_evaluation}. The detailed prompts of Counter-entity and Counter-Semantic in~\ref{tab:entity_replacement_framework} and \ref{tab:counterfactual_evidence_prompt}. The prompts of baselines are shown in Table~\ref{tab:ImplicitSCR and ExplicitSCR}, Table~\ref{tab:InternalEval and ContextEval} and Table~\ref{tab:talr_framework}.

\section{LLM Usage Claim} 
\label{app:H}
In this paper, we used GPT-4o to assist counterfactual evidence construction under controlled prompting. All generated counterfactuals were filtered and partially audited by human annotators.

Additionally, LLMs are utilized exclusively for the purpose of aiding and polishing writing. Their application is strictly confined to improving linguistic clarity, coherence, grammar, and style within
textual content. No additional functionalities are incorporated.

\begin{table*}[htbp]
\centering
\footnotesize
\renewcommand{\arraystretch}{1.4} % 增加行高，容纳长文本
\setlength{\tabcolsep}{3pt}    % 缩小列间距以平衡列宽

\begin{tabular}{>{\RaggedRight}p{3.8cm}|c|>{\RaggedRight}p{5.1cm}|>{\RaggedRight}p{5.1cm}}
\toprule
\textbf{Claim} ($\mathbf{c}$) & \textbf{Label} & \textbf{Evidence ($\mathbf{e}$)} & \textbf{Constructed Evidence  with counterfactual ($\mathbf{\bar{e}}$)} \\
\midrule

% 修正后的分界行
\multicolumn{4}{c}{\textbf{--- crawled new claims ---}} \\
\midrule

% 第三行：2025 乌克兰
On April 6 2025: Russian invasion of Ukraine. 2025 Sumy Oblast incursion Russian troops reportedly capture the village of Basivka in Sumy Oblast, Ukraine. (Reuters) Kyiv strikes A Russian airstrike in Darnytskyi District, Kyiv, Ukraine, kills one person and injures three others. (CTV News) Kryvyi Rih strikes The death toll from Friday's missile strike on Kryvyi Rih, Ukraine, rises to 20 deaths, including several children, and 75 injuries. (CTV News) & True & 
... In an effort to counter \textcolor{textblue}{Ukraine's 2024 offensive} in \textcolor{textblue}{Kursk}, in early \textcolor{textblue}{2025} Russian forces launched a cross-border offensive from \textcolor{textblue}{Russia} into Ukraine's adjacent \textcolor{textblue}{Sumy Oblast}. ... & 
... In an effort to counter \textcolor{textred}{Poland's "2026 offensive"} in \textcolor{textred}{Voronezh}, in early \textcolor{textred}{2026} Russian forces launched a "counter-terrorism operation" from \textcolor{textred}{Germany} into Poland's adjacent \textcolor{textred}{Cherkasy Oblast}. ... \\ \midrule

% 第四行：2025 红海
On April 6 2025: Red Sea crisis March–April 2025 United States attacks in Yemen. Between four and 70 people are killed and at least 16 others are injured in overnight U.S. airstrikes targeting Houthi forces in Saada, Yemen. (CTV News) & True & 
... In \textcolor{textblue}{March 2025}, the United States launched a large campaign of air and naval strikes against Houthi targets in \textcolor{textblue}{Yemen}. Codenamed Operation \textcolor{textblue}{Rough Rider}, it has been the largest U.S. military operation in the Middle East of President \textcolor{textblue}{Donald Trump's second term} ... & 
... In \textcolor{textred}{March 2026}, the United States launched a large campaign of air and naval strikes against Houthi targets in \textcolor{textred}{Oman}. Codenamed Operation \textcolor{textred}{Iron Stallion}, it has been the largest U.S. military operation in the Middle East of President \textcolor{textred}{Joe Biden's second term} ... \\

\bottomrule
\end{tabular}
\caption{Examples of external evidence with counterfactual (Counter-entity).}
\label{Table: examples_db_fact_conf}
\end{table*}

\begin{table*}[htbp]
\centering
\footnotesize
\renewcommand{\arraystretch}{1.8} % 增加行高，容纳大量文字
\setlength{\tabcolsep}{5pt}

\begin{tabular}{p{3cm} | c | p{5cm} | p{5cm}}
\toprule
\textbf{Claim} ($\mathbf{c}$) & \textbf{Label} & \textbf{Evidence ($\mathbf{e}$)} & \textbf{Constructed Evidence with counterfactual ($\mathbf{\bar{e}}$)} \\
\midrule

% Case 1: Tax Loopholes (红色标记)
\RaggedRight federal tax code loopholes giving incentives companies shipping jobs overseas & True & 
... I also want to close those loopholes that are giving incentives for companies that are shipping jobs overseas. Obama said Wednesday right now you can actually take a deduction for moving a plant overseas... & 
\textcolor{purple}{For firms choosing to remain in the United States, the current tax code provides significant incentives for companies that invest domestically. Instead of creating disincentives for offshoring, the tax structure encourages businesses to expand their operations within the U.S.} by offering substantial tax credits for job creation and capital investment. \\ \hline

% Case 2: Mitt Romney (紫色标记)
\RaggedRight Mitt Romney gave government health care bankrupting state Massachusetts & False & 
It has been nearly five years since Massachusetts gov mitt Romney signed the state's landmark health care law... the overhaul is largely seen as a blueprint for the sweeping federal legislation. & 
... \textcolor{purple}{However, the law has faced significant backlash, with many residents expressing dissatisfaction over rising premiums and limited access to care. Critics argue that the law has not delivered on its promises... Brian Rosman has since distanced himself from the initial celebration.} \\
% \hline

% % Case 4: President Bush (橙色标记)
% \RaggedRight 2003 president Mrs. Bush helped hand Christmas presents children inmates & True & 
% In 2003 president and Mrs. Bush helped hand out Christmas presents to children of inmates. status true... angel tree our prison fellowship program. & 
% In \textcolor{purple}{2005 governor Mrs. Clinton give holiday gifts to teenagers of prisoners. condition false... star branch our rehabilitation outreach initiative for refugee families} is one of the great unheralded intern. \\

\bottomrule
\end{tabular}
\caption{Examples of external evidence with counterfactual (Counter-semantic).}
\label{tab:counter_semantic}
\end{table*}

\begin{table*}[htbp]
\centering
\small
\renewcommand{\arraystretch}{1.5}
% 修复：定义 4 列，总宽度约 16.5cm，Evidence 给予最大空间
\begin{tabular}{>{\RaggedRight}p{5cm} | >{\RaggedRight}p{5cm} | c | >{\RaggedRight\arraybackslash}p{3cm}}
\toprule
\textbf{Claim} & \textbf{Evidence} & \textbf{Time} & \textbf{Outdated for LLMs}\\
\midrule

% Row 1
July 31, 2025. The military junta of Myanmar formally ends the country's four-year-long state of emergency and declares a December 2025 election for the country's new head of government and legislative members. & 
...Myanmar's military government plans to hold a general election for elected seats in the Amyotha Hluttaw and the Pyithu Hluttaw of the Assembly of the Union, currently dissolved, on a date in December 2025 to be determined. The planned election would be the first after the 2021 military coup d'état... & 
July 31, 2025 & 
Deepseek-v3, GPT-4o-mini, Llama3-8B, Phi-4, Gemini-2.5-flash, Mistral-7B \\ \hline

% Row 2
2025 FIFA Club World Cup. In association football, English club Chelsea F.C. win the FIFA Club World Cup for the second time, defeating French club Paris Saint-Germain 3–0 in the final. Chelsea winger Cole Palmer is awarded the tournament's Golden Ball. (FIFA) &
The 2025 FIFA Club World Cup final was the final match of the 2025 FIFA Club World Cup, the 21st edition of the premier competition for men's club soccer teams organized by FIFA. The match was played at MetLife Stadium at the Meadowlands Sports Complex in East Rutherford, New Jersey, near New York City, on July 13, 2025. It was contested between English club Chelsea and French club Paris Saint-Germain. & 
July 13, 2025 & 
Deepseek-v3, GPT-4o-mini, Llama3-8B, Phi-4, Gemini-2.5-flash, Mistral-7B \\ \hline

% Row 3
2024 Maltese presidential election. Myriam Spiteri Debono is sworn in as President of Malta, succeeding George Vella and becoming the third woman to hold the office. (Times of Malta) & 
The 2024 Maltese presidential election took place on 27 March 2024. Members of the Parliament of Malta voted in an indirect election to elect the next President of Malta with former parliament speaker Myriam Spiteri Debono being the only nominee & 
April 4, 2024 & 
Mistral-7B \\ \hline

% Row 4
Arab–Israeli conflict. Gaza war. Gaza war hostage crisis. Hamas releases a propaganda video showing Israeli American hostage Edan Alexander. (CBS News). An Israeli airstrike on a car in the Gaza Strip kills five people, including employees of World Central Kitchen. (AP). Israel–Hezbollah conflict. Two people are killed and six others are injured in three airstrikes by Israel in southern Lebanon. &
On 27 November 2024, a ceasefire agreement was signed by Israel, Lebanon, and five mediating countries. The agreement mandates a 60-day halt to hostilities, during which Israel must withdraw its forces from Southern Lebanon, and Hezbollah must withdraw its forces to north of the Litani River. Since the ceasefire went into effect, Lebanese sources claim Israeli attacks on Lebanon killed at least 83 civilians. & 
Nov 30, 2024 & 
GPT-4o-mini, Llama3-8B, Mistral-7B \\ 
% \hline

% % Row 5
% July 31, 2025. The military junta of Myanmar formally ends the country's four-year-long state of emergency and declares a December 2025 election for the country's new head of government and legislative members. & 
% ... Myanmar's military government plans to hold a general election for elected seats in the Amyotha Hluttaw and the Pyithu Hluttaw of the Assembly of the Union, currently dissolved, on a date in December 2025 to be determined. The planned election would be the first after the 2021 military coup d'état ... & 
% July 31, 2025 & 
% GPT-4o-mini, Llama3-8B, Phi-4, Deepseek-v3, Gemini-2.5-flash, Mistral-7B \\
\bottomrule
\end{tabular}
\caption{We present examples of crawled events with evidence from our benchmarks. This information is new for some LLMs, indicating that their parametric knowledge is outdated.}
\label{App: examples_db2_table}
\end{table*}

\begin{table*}[h]
\small
\centering
\begin{tabular}{|p{0.95\textwidth}|}

\hline
\rowcolor{gray}  \multicolumn{1}{|c|}{\textbf{Parametric or External Knowledge}} \\
\hline\hline
\textbf{Parametric Knowledge Assessment (Without Evidence)} \\
Based solely on your internal knowledge, determine whether the claim should be supported or refuted. Provide only "support" or "refute" as the final answer without any explanation.\\
\\
\textbf{Task:}\\
Claim: \{claim\}\\
Answer:\\
\hline
\end{tabular}

\begin{tabular}{|p{0.95\textwidth}|}
\hline
\textbf{External Knowledge Integration (With Evidence)} \\
Based on the provided information and your internal knowledge, determine whether the claim should be supported or refuted. Provide only "support" or "refute" as the final answer without any explanation.\\
\\
\textbf{Task:}\\
Claim: \{claim\}\\
Information: \{evidence\}\\
Answer:\\
\hline
\end{tabular}
\caption{The basic prompts of evaluation.}
\label{tab:basic_evaluation}

\end{table*}

\begin{table*}[h]

\centering
\begin{tabular}{|p{0.95\textwidth}|}
\hline
\rowcolor{gray}  \multicolumn{1}{|c|}{\textbf{Counterfactual Evidence Generation}} \\
\hline\hline
\textbf{True Claim Refutation Prompt} \\
Given a claim labeled as true, generate realistic but fabricated evidence to refute it. Create convincing fake content that contradicts the true claim.\\
\\
\textbf{Task:}\\
Claim: \{claim\} (label: true)\\
Evidence:\\
\hline
\end{tabular}

\begin{tabular}{|p{0.95\textwidth}|}
\hline
\textbf{False Claim Support Prompt} \\
Given a claim labeled as false, generate realistic but fabricated evidence to support it. Create convincing fake content that validates the false claim.\\
\\
\textbf{Task:}\\
Claim: \{claim\} (label: false)\\
Evidence:\\
\hline
\end{tabular}
\caption{Prompts of counterfactual semantic evidence generation framework. We use the counter-label of claim to set the prompts.}
\label{tab:counterfactual_evidence_prompt}
\end{table*}

\begin{table*}[h]

\centering
\begin{tabular}{|p{0.95\textwidth}|}
\hline
\rowcolor{gray}  \multicolumn{1}{|c|}{\textbf{Entity Substitution}} \\
\hline\hline
\textbf{Step 1: Entity Extraction Prompt} \\
Extract the entities from the following evidence that directly influence the judgment of the claim.\\
\\
\textbf{Example:}\\
Claim: On April 6 2025: Red Sea crisis March–April 2025 United States attacks in Yemen. Between four and 70 people are killed and at least 16 others are injured in overnight U.S. airstrikes targeting Houthi forces in Saada, Yemen.\\
Evidence: ... In March 2025, the United States launched a large campaign of air and naval strikes against Houthi targets in Yemen. Code-
named Operation Rough Rider, it has been the largest U.S. military operation in the Middle East of President Donald Trump’s second term ...\\
Entities: ...March 2025, Yemen, Rough Rider, Donald Trump’s second term...\\
\\
\textbf{Task:}\\
Claim: \{claim\}\\
Evidence: \{evidence\}\\
\hline
\end{tabular}

\begin{tabular}{|p{0.95\textwidth}|}
\hline
\textbf{Step 2: Replaced Entity Generation Prompt} \\
Given the entity, generate a similar type but a different entity. Only output the new entity.\\
\\
\textbf{Example:}\\
Entity: ...{March 2025}, {Yemen}, {Rough Ride}, {Donald Trump’s second term}...\\
New Entity: ...{March 2026}, {Oman}, {Iron Stallio}, {Joe Biden’s second term}\\
\\
\textbf{Task:}\\
Entity: \{ent\}\\
\hline
\end{tabular}
\caption{The prompts of Counter-entity construction. We let LLM generate the replaced entity one by one.}
\label{tab:entity_replacement_framework}

\end{table*}

\begin{table*}[h]

\centering
\begin{tabular}{|p{0.95\textwidth}|}
\hline
\rowcolor{gray} \multicolumn{1}{|c|}{\textbf{ImplicitSCR}} \\
\hline\hline
\textbf{Prompt:} You will be given a question and an evidence. The evidence may not be trustworthy. Use your judgment to assess the reliability of the evidence. Then, based on both your assessment and your own knowledge, provide the best possible answer.\\
\textbf{Question:} [question]\\
\textbf{Evidence:} [evidence]\\
\textbf{Answer:} \\
\hline
\end{tabular}

% \vspace{-1cm}

\begin{tabular}{|p{0.95\textwidth}|}
\hline
\rowcolor{gray}  \multicolumn{1}{|c|}{\textbf{ ExplicitSCR}} \\
\hline\hline
\textbf{Prompt:} Task Overview: You will be given a question along with your internal answer, evidence that may contain either true or false information, and the evidence's answer to the same question. Your task is to evaluate the reliability of the evidence and determine whether the evidence is deceptive or not.\\
Steps:\\
1. Internal Reasoning: Reflect on how you arrived at your internal answer using your own knowledge. Break down your reasoning process and assess the confidence level of your original answer, explaining why you believe your answer is correct.\\
2. Evidence Evaluation: Analyze the evidence and cross-reference the information provided with the known facts you used to form your internal answer. Determine whether the evidence contains deceptive or unreliable information, considering possible contradictions or inconsistencies.\\
3. Final Judgment: Based on your analysis, decide which answer (your internal answer or the evidence's answer) is more likely to be correct. Clearly state your final answer.\\
\textbf{Question:} \{question\}\\
\textbf{Your answer:} \{internal answer\}\\
\textbf{The evidence to judge:} \{evidence\}\\
\textbf{The evidence answer:} \{evidence answer\}\\
Please provide a detailed reasoning process, followed by your final judgment. Ensure the last line of your response contains only the final answer without any additional explanation or details. \\
\hline
\end{tabular}
\caption{The prompt structures of ImplicitSCR and ExplicitSCR}
\label{tab:ImplicitSCR and ExplicitSCR}

\end{table*}

\begin{table*}[h]

\centering
\begin{tabular}{|p{0.95\textwidth}|}
\hline
\rowcolor{gray}  \multicolumn{1}{|c|}{\textbf{InternalEval}} \\
\hline\hline
\textbf{Prompt:} Your task is to evaluate the model's response to a question. You will be provided with a question, the model's answer. Your job is to determine whether the model's answer is true or false.\\
\textbf{Question:} [question]\\
\textbf{Model Answer:} [model answer]\\
\textbf{Is the model's answer true or false?}\\
Return "True" if the model's answer is correct, and "False" if the model's answer is incorrect. \\
\hline
\end{tabular}

% \vspace{1cm}

\begin{tabular}{|p{0.95\textwidth}|}
\hline
\rowcolor{gray}  \multicolumn{1}{|c|}{\textbf{ContextEval}} \\
\hline\hline
\textbf{Prompt:} You will be given a question and evidence that answers the question. Your task is to evaluate whether the evidence provides a correct answer to the question. If the evidence's answer is correct, return "True"; otherwise, return "False".\\
\textbf{Question:} [question]\\
\textbf{Evidence:} [doc]\\
\textbf{Is the evidence correct?}\\
Return "True" if the evidence's answer is correct, and "False" if the evidence's answer is incorrect. \\
\hline
\end{tabular}
\caption{The prompt structures of InternalEval and ContextEval}
\label{tab:InternalEval and ContextEval}

\end{table*}

\begin{table*}[h]

\centering
\begin{tabular}{|p{0.95\textwidth}|}
\hline
\rowcolor{gray}  \multicolumn{1}{|c|}{\textbf{TACS-LR}} \\
\hline\hline
\textbf{Step 1: Evidence Filtering Prompt} \\
You will be given an Evidence and a question. You need to remove the sentence that you think is not correct. 
You can only do removal and you can not add any new information or change the existing information. Only return the filtered evidence as your output.\\
\\
\textbf{Example:}\\
Evidence: The Eiffel Tower is located in Paris, France. It is the tallest structure in Paris. The Eiffel Tower was built in the 19th century and is made of wood.\\
Question: Where is the Eiffel Tower located?\\
Filtered Evidence: The Eiffel Tower is located in Paris, France. It is the tallest structure in Paris. The Eiffel Tower was built in the 19th century.\\
\\
\textbf{Task:}\\
Question: \{question\}\\
Evidence: \{evidence\}\\
\hline
\end{tabular}

\begin{tabular}{|p{0.95\textwidth}|}
\hline
\textbf{Step 2: Final Answer Generation Prompt} \\
Your task is to answer the "Question" based on the provided "Evidence".\\
\\
\textbf{Example:}\\
Question: Is the claim 'The capital of France is Paris' true or false? Do not generate the process, just answer "true" or "false" only!\\
Evidence: Paris has been the administrative, political, and cultural capital of France since 987 AD when the Capetian dynasty established their power base there, a status that has remained uninterrupted for over a thousand years.\\
Answer: true\\
\\
\textbf{Task:}\\
Question: \{question\}\\
Evidence: \{filtered evidence\}\\
\hline
\end{tabular}
\caption{The prompt structure of TACS-LR}
\label{tab:talr_framework}

\end{table*}

\end{document}